\pdfoutput=1
\documentclass[10pt,twocolumn,letterpaper]{article}

\usepackage[pagenumbers]{cvpr} %

\usepackage{cuted} %

\usepackage{currfile} %

\usepackage{caption} %

\usepackage{array}
\usepackage{multirow}
\usepackage{tabularx}
\usepackage{colortbl}
\newcommand{\dashsep}[1]{\arrayrulecolor{gray!55}\cmidrule[0.3pt](lr){#1}\arrayrulecolor{black}}
\newcommand{\vlabel}[1]{\rotatebox[origin=c]{90}{\colorbox{tabgray}{\scriptsize #1}}}
\newcommand{\noagent}[1]{\,{\color{gray}\scriptsize(#1)}}
\usepackage{pifont}
\usepackage{enumitem}
\usepackage{listings}
\setlist{nosep,leftmargin=*}
\newcolumntype{Y}{>{\centering\arraybackslash}X}
\newcolumntype{L}{>{\raggedright\arraybackslash}X}

\definecolor{tabgray}{HTML}{F4F6F8}
\definecolor{tabblue}{HTML}{EBF3FA}
\definecolor{tabgreen}{HTML}{EAF6EE}

\newcommand{\mascot}[1]{}
\newcommand{\method}{LatentParse}

\definecolor{cvprblue}{rgb}{0.21,0.49,0.74}
\usepackage[pagebackref,breaklinks,colorlinks,allcolors=cvprblue]{hyperref}
\usepackage[capitalize]{cleveref}

\def\confName{CVPR}
\def\confYear{2027}

\title{Parsing Visual Latents for Multimodal Spatial Reasoning}

\author{%
Zhuoying Yang$^{1*}$\quad
Yichen Guo$^{2*}$\quad
Kathy Jia$^{1*}$\quad
Zichang Su$^{1*}$\quad
Ayesha Kashif$^{1}$\\
Nange Hai$^{1}$\quad
Jeffrey Ming Han Li$^{3}$\quad
Zhewen He$^{4}$\quad
Silvio Kempf$^{1}$\quad
Ni Jiachen$^{5}$\\
Junjie Yang$^{6}$\quad
Muhammad Hamza$^{3}$\quad
Zakaria El Ouahidi$^{7,1}$\quad
Gabriel Azpitarte$^{1}$\\
Manling Li$^{8\dagger}$\quad
Yi Fang$^{4\dagger}$\quad
Shuhong Liu$^{9\dagger}$\quad
Mengyu Wang$^{1\dagger}$\\[6pt]
{\small $^1$Harvard University\quad
$^2$Nanyang Technological University\quad
$^3$University of Toronto\quad
$^4$New York University\quad
$^5$Tufts University}\\
{\small $^6$University of Maryland, College Park\quad
$^7$Centrale Sup\'elec\quad
$^8$Northwestern University\quad
$^9$The University of Tokyo}\\[2pt]
{\small $^*$Co-first authors\qquad $^\dagger$Co-senior authors}%
}

\begin{document}
\maketitle

\begin{abstract}

Spatial reasoning across multiple views requires inferring the 3D relations between objects and the camera from 2D observations, which depends on cross-view correspondence and scene geometry that vision-language models (VLMs) do not reliably extract. Pretrained vision encoders already capture this information in their latent features, with CLIP linking objects to image regions, DINOv3 matching semantics across views, and VGGT encoding multi-view geometry. However, VLMs do not benefit from these features when we present them directly in context as grayscale feature images. We therefore propose \method, a training-free framework that parses latent features into explicit spatial evidence for VLMs. In \method, an analyst VLM probes the features of all three encoders with predefined tools that locate objects, track them across views, and measure motion between views. It then distills the results into a concise text report, which is provided together with the images to the VLM to answer the question. We evaluate \method\ on MMSI-Bench, MV-RoboBench and SPAR-Bench, where it improves all five VLMs over the photo-only baseline by 0.7--2.4 percentage points on average. On MMSI-Bench, it raises Qwen3.8-27B from 34.4\% to 39.5\%, with the largest improvements of 19.2 percentage points on the camera-related questions. These results suggest that latent features become useful for spatial reasoning once they are parsed into relation evidence, without training. Code is available at \url{https://github.com/Harvard-AI-and-Robotics-Lab/LatentParse}.

\end{abstract}

\section{Introduction}
\label{sec:intro}

\begin{figure}[t]
    \centering
    \includegraphics[width=\columnwidth]{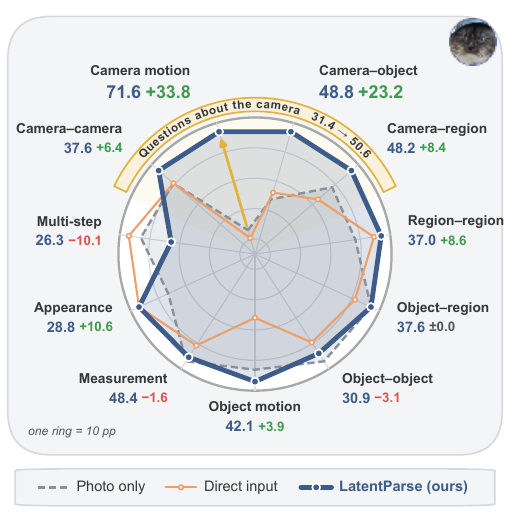}
    \caption{\mascot\ \textbf{\method\ helps most on questions about the camera.} Qwen3.8-27B on the 11 MMSI-Bench question types. Each label gives \method's accuracy and its change from photo only. On the four camera types, marked orange, accuracy rises from $31.4\%$ to $50.6\%$ on the 334 questions with a paired photo-only answer.}
    \label{fig:radar_types}
\end{figure}
\begin{figure*}[t]
    \centering
    \IfFileExists{fig/fig_evidence_mmsi.pdf}{%
        \includegraphics[width=0.98\textwidth]{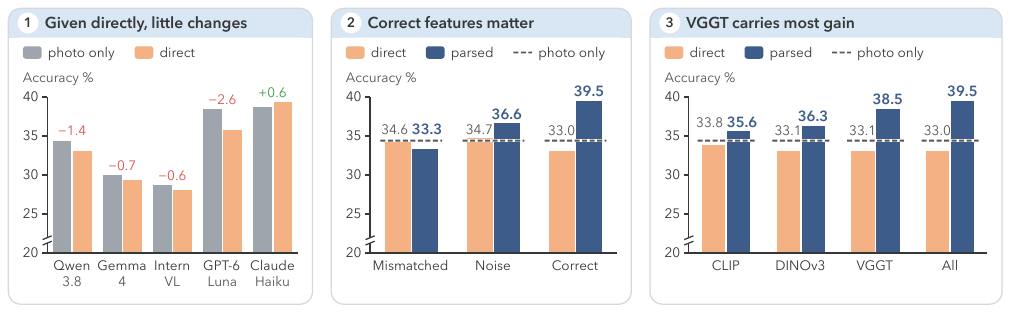}%
    }{\includegraphics[width=0.98\textwidth]{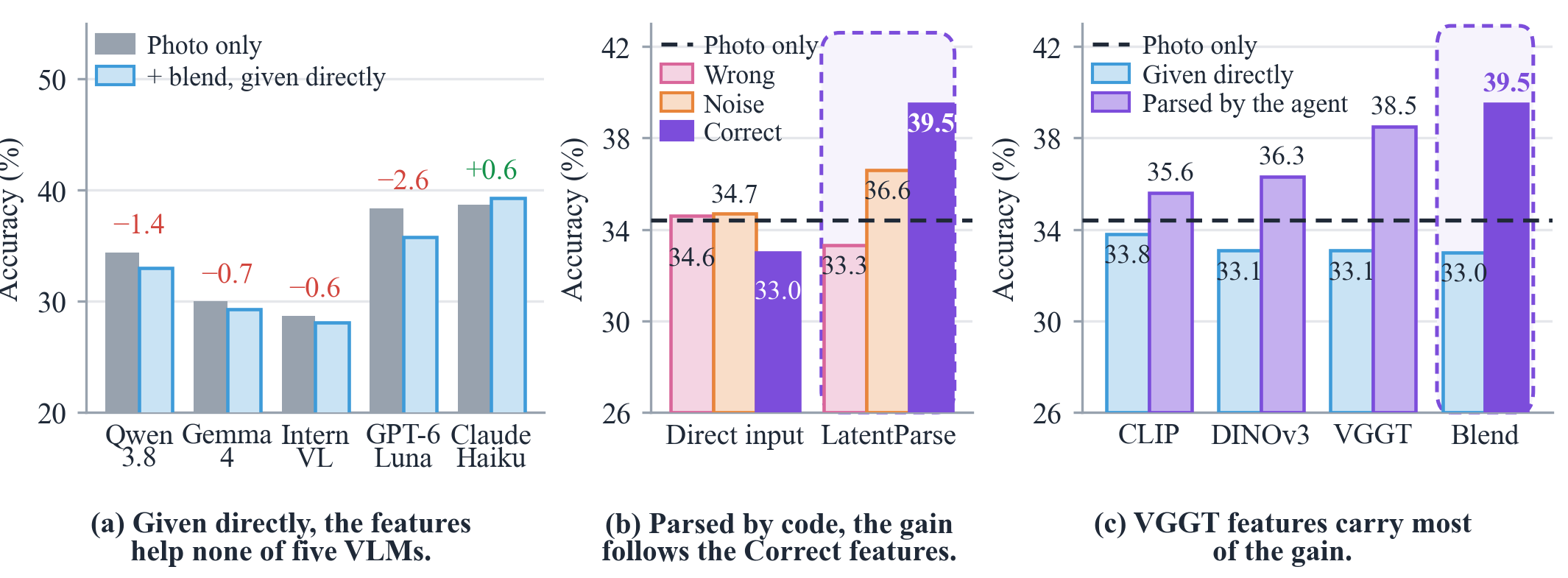}}
    \caption{\mascot\ \textbf{Parsing latent features improves accuracy over the tested feature images.} All panels use the 1,000 MMSI-Bench questions. \textbf{(1)} Given directly, the features of all three encoders (Multi-Latent) change five VLMs by $-2.6$ to $+0.6$~pp. \textbf{(2)} For Qwen3.8-27B, \method\ scores highest with the Correct features, while direct input stays near photo only. \textbf{(3)} Each encoder's features and all three, given directly or parsed by the analyst. With all three encoders the analyst is \method; with one encoder it is the earlier single-encoder analyst of \Cref{sec:exp_setup}. VGGT alone reaches $38.5\%$, against $39.5\%$ with all three.}
    \label{fig:evidence}
\end{figure*}

Vision-language models (VLMs) have advanced rapidly in recognizing and describing visual content~\cite{liu2024llavanext,tong2024cambrian}, yet they still struggle to reason about the 3D layout of a scene and the motion of the camera~\cite{fu2024blink,ma20243dsrbench,yang2025mmsibench,yang2024thinking}. Answering these questions requires inferring 3D relations such as the relative position, distance, and orientation between objects and viewpoints~\cite{kamath2023whatsup,chen2024spatialvlm,yang2024thinking}. These relations are often underdetermined by a single image and must be recovered from cross-view correspondence and scene geometry, neither of which current VLMs explicitly model.

Pretrained vision encoders offer a complementary source of this information. In particular, CLIP~\cite{radford2021clip}, DINOv3~\cite{simeoni2025dinov3} and VGGT~\cite{wang2025vggt} are trained with image–text alignment, self-supervision and multi-view 3D supervision, respectively. Preliminary probing studies reveal that these features encode rich semantic and geometric information~\cite{elbanani2024probe3d,man2024lexicon3d,chen2025feat2gs,zhang2022depthclip}, making them a reliable prior for spatial reasoning. Recent studies exploit this prior by training the VLMs to interpret encoder features. For example, Spatial-MLLM~\cite{wu2025spatialmllm}, VG-LLM~\cite{zheng2025vgllm} and VLM-3R~\cite{fan2025vlm3r} connect visual encoders to the language model through learned projectors or fusion modules and fine-tune on spatial instruction data. Such training must be repeated for each encoder–VLM pair, which is computationally expensive and cannot be applied to closed-source VLMs. Training-free methods, including ViperGPT~\cite{suris2023vipergpt}, SpatialPIN~\cite{ma2024spatialpin} and SandboxVLM~\cite{liu2026sandboxvlm}, instead support reasoning with decoded outputs such as depth maps, reconstructions or 3D boxes. These outputs are fixed before the question is known and may discard the cross-view correspondence and region similarity that the latent features encode and many spatial questions require.

A natural alternative is to use the latent features themselves as context, which require neither training nor decoding and leave the VLM free to decide what to measure for each question. However, as \Cref{fig:evidence} shows, presenting these latent features to a VLM naively as image input brings little benefit, likely because the information they carry lies in the relations between patches rather than in their appearance. To exploit these relations and realize these benefits, we propose LatentParse, a training-free framework that parses the relations between patches into explicit evidence and summarizes it in a text report that an arbitrary VLM can use in context, as illustrated in \Cref{fig:pipeline}.

\method\ operates on the last-layer features of CLIP~\cite{radford2021clip}, DINOv3~\cite{simeoni2025dinov3} and VGGT~\cite{wang2025vggt}, which respectively ground object names in image regions, compare regions within and across views, and relate the views geometrically. On these features, a library of tools locates named objects, tracks them across views, measures content motion, and rates each measurement by consensus. An analyst VLM that never sees the photos calls these tools in a Python sandbox~\cite{lee2026metaharness,wei2026selfevolvingharness} and writes a report. The same VLM then answers the question in a fresh context from the photos and this report. Keeping the analyst blind to the photos is intended to focus its report on tool measurements rather than direct photo inspection.

We evaluate \method\ with five open and closed VLMs on three spatial reasoning benchmarks~\cite{yang2025mmsibench,feng2026mvrobobench,zhang2025spar}. Across the benchmarks, our method raises the mean accuracy of every VLM over the photo-only baseline by 0.7 to 2.4 pp, whereas naive input changes accuracy on MMSI-Bench by only $-2.6$ to $+0.6$~pp. On MMSI-Bench, \method\ improves Qwen3.8-27B from 34.4\% to 39.5\%, with the largest gains on camera-related questions, which rise from 31.4\% to 50.6\% as shown in \Cref{fig:radar_types}. Using mismatched features from a different scene lowers accuracy to 33.3\%, which indicates that the gain depends on features that match the scene.

In summary, our contributions are as follows:
\begin{itemize}
\item We propose \method, a training-free framework that enables frozen VLMs to exploit the latent features of pretrained vision encoders for spatial reasoning by parsing these features into relation evidence through a photo-blind analyst and a textual report.
\item We design a library of latent-space tools for object localization, cross-view tracking, and motion estimation, which rely mainly on basis-invariant comparisons within each encoder and rate measurements by cross-encoder consensus, without any learned projector.
\item We evaluate five open- and closed-source VLMs on three benchmarks, where \method\ improves the mean accuracy of every VLM over the photo-only baseline by 0.7 to 2.4 pp whereas the tested grayscale feature images yield no consistent gain. For Qwen3.8-27B on MMSI-Bench, the gain disappears with mismatched features, concentrates on camera questions and turns into a loss on multi-step reasoning.
\end{itemize}

\section{Related Work}
\label{sec:related}

\subsection{Spatial Reasoning in Vision Language Models}
\label{sec:related_spatial}
\label{sec:related_interfaces}
A growing body of benchmarks evaluates whether VLMs can infer scene geometry from a single image and relate observations across multiple views.
3DSRBench~\cite{ma20243dsrbench} focuses on depth order, orientation and object relations within a single image, whereas MMSI-Bench~\cite{yang2025mmsibench}, MindCube~\cite{wang2026mindcube} and MV-RoboBench~\cite{feng2026mvrobobench} require reasoning about camera motion, hypothetical movements and object correspondence across views.
SPAR-Bench~\cite{zhang2025spar} and VSI-Bench~\cite{yang2024thinking} further extend this evaluation to indoor scenes and video.
To improve performance on these benchmarks, most existing methods train the VLM itself.
SpatialVLM~\cite{chen2024spatialvlm} is trained on synthetic spatial question-answer pairs, SpatialRGPT~\cite{cheng2024spatialrgpt} learns regional representations from 3D scene graphs, and GASP~\cite{yeh2026beyond3dvqa} derives supervision from correspondence and depth consistency without relying on 3D VQA data.
Another line of work injects features from additional encoders through a trained connector, in the spirit of the pretrained projector in LLaVA-NeXT~\cite{liu2024llavanext}.
Cambrian-1~\cite{tong2024cambrian} aggregates features from multiple encoders with a learned Spatial Vision Aggregator, while Spatial-MLLM~\cite{wu2025spatialmllm}, VG-LLM~\cite{zheng2025vgllm} and VLM-3R~\cite{fan2025vlm3r} incorporate geometry features through learned connectors or fusion modules.
GeoReader~\cite{li2026georeader} trains Readers that encode predicted depth and camera poses as continuous tokens, and subsequently fine-tunes the VLM for spatial QA with the Readers frozen.
Since such training has to be repeated for every encoder-VLM pair, these approaches are costly to extend and cannot be applied to closed-source VLMs.
In contrast, LatentParse requires neither a learned connector nor any VLM update.

\subsection{Spatial Information in Vision Encoder Latents}
\label{sec:related_encoders}
Probing studies investigate how much spatial information can be recovered from frozen vision features.
Probe3D~\cite{elbanani2024probe3d} evaluates depth estimation, surface-normal prediction and cross-view correspondence, including on DINOv2~\cite{oquab2023dinov2} features, and Lexicon3D~\cite{man2024lexicon3d} extends this analysis to 3D grounding, segmentation and registration.
Feat2GS~\cite{chen2025feat2gs} assesses the recovered geometry through novel-view synthesis, and DepthCLIP~\cite{zhang2022depthclip} shows that coarse depth can be read from CLIP~\cite{radford2021clip} even without fitting a dedicated probe.
These findings establish that rich spatial information resides in encoder features, yet leave open whether a separate VLM can exploit it to answer spatial questions.
A related line of research performs reasoning directly in latent space.
Coconut~\cite{hao2024coconut} reasons in a continuous latent space, LatentMAS~\cite{zou2025latentmas} exchanges hidden states among collaborating agents, and Latent Sketchpad~\cite{zhang2025latentsketchpad} lets a model sketch within its own latent space.
These methods, however, operate on the model's internal hidden states, which the model is trained to produce and interpret.
Our work instead draws on the patch features of external encoders, namely CLIP, DINOv3~\cite{simeoni2025dinov3} and VGGT~\cite{wang2025vggt}, whose channels bear no correspondence to the VLM's visual tokens.
We therefore compare presenting these features directly as images with parsing the relations among their patches into textual evidence.

\subsection{Training-Free Visual Tool Use}
\label{sec:related_tools}
\label{sec:related_harness}
Training-free methods equip a model with vision tools rather than modifying its weights.
VisProg~\cite{gupta2023visprog} and ViperGPT~\cite{suris2023vipergpt} formulate visual reasoning as programs that invoke perception modules, and Visual Sketchpad~\cite{hu2024visualsketchpad} produces intermediate drawings for the VLM to inspect.
OpenVLHarness~\cite{zhou2026openvlharness} retains tool outputs in a multimodal memory across steps, while MAIA~\cite{shaham2024maia} executes code to interpret model components rather than the scene itself.
Closer to our setting, SpatialPIN~\cite{ma2024spatialpin} supplies explicit 3D priors for spatial reasoning, and SandboxVLM~\cite{liu2026sandboxvlm} renders scene abstractions from 3D bounding boxes.
Perception Programs~\cite{janjua2026perceptionprograms} condenses the outputs of depth, flow and correspondence tools into compact text, and Coarse Correspondences~\cite{liu2025coarse} overlays objects tracked across views directly on the images.
Such tool use is typically organized by an \emph{agent harness}, which governs the tools available to a model, how each call is executed and what information persists between steps~\cite{wong2025confucius,yao2026harnessbench}.
Recent work adapts harnesses by refining prompts and workflows~\cite{jiang2026harnessevolve,lee2026metaharness,lin2026ahe,pan2026rho,wei2026selfevolvingharness,xu2026harness2,zhang2026selfharness}, updating memory during deployment~\cite{karten2026continualharness,xiao2026reharness}, improving execution efficiency~\cite{liu2026solpi}, or training the model on harness edits and demonstrations~\cite{shao2026harnessr1,ye2026harnesszero}.
LatentParse adopts a fixed code-execution loop of this kind, in which a photo-blind analyst applies predefined tools to the patch features of off-the-shelf encoders and summarizes its findings in a report that the answering VLM reads alongside the photos.
Unlike prior methods that operate on images or on the decoded outputs of tools and task heads, our analyst measures relations within each encoder's latent features, and rates each finding by cross-encoder consensus.

\section{Method}
\label{sec:method}
\begin{figure*}[t]
    \centering
    \includegraphics[width=0.98\textwidth]{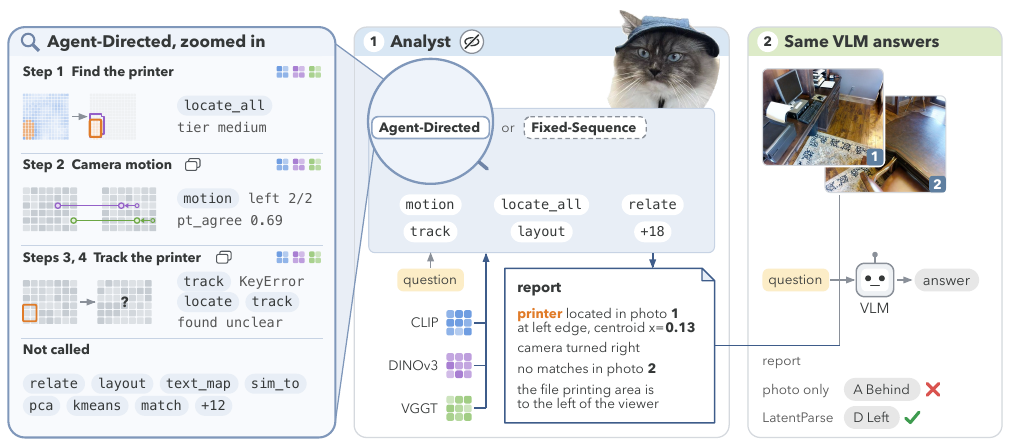}
    \caption{\mascot\ \textbf{Overview of \method.} The photo-blind analyst calls predefined tools in up to four steps on the latent features of CLIP, DINOv3 and VGGT, zoomed in at left for one MMSI-Bench question, and writes a report, shown abridged. A separate call to the same VLM in a fresh context answers from the photos and the report.}
    \label{fig:pipeline}
\end{figure*}

A spatial question $q$ comes with answer options and $V \ge 1$ views $\{I_v\}_{v=1}^V$. We want a VLM to benefit, without training, from the per-patch latent features $F^{(k)} \in \mathbb{R}^{V \times h \times w \times d_k}$ of a vision encoder $\mathcal{E}_k$. As shown in \Cref{fig:pipeline}, \method{} first extracts these features from CLIP, DINOv3 and VGGT. An analyst VLM then parses them with tools and writes a textual report, and the same VLM answers the question from the photos and this report in a fresh context. \Cref{sec:method_extract} describes the latent features and why direct input fails, \Cref{sec:method_tools} introduces the tools that measure them, and \Cref{sec:method_analyst} presents the analyst and the answering step.

\subsection{Latent Features as Context}
\label{sec:method_extract}

\paragraph{Three complementary encoders.} We draw latent features from three pretrained encoders whose training objectives capture different aspects of a scene. CLIP~\cite{radford2021clip} is trained with image--text alignment and links object names to image regions. DINOv3~\cite{simeoni2025dinov3} is trained with self-supervision and produces features that match semantically similar regions within and across views. VGGT~\cite{wang2025vggt} is trained with multi-view 3D supervision and encodes how the views relate geometrically. From each encoder we take the patch tokens of the last layer and never use the depth, point or camera heads of VGGT. These latent features retain information that decoded outputs such as depth maps or 3D boxes~\cite{ma2024spatialpin,liu2026sandboxvlm} discard, and leave the choice of what to measure to the analyst. \Cref{tab:app_layers} compares other layers.

\paragraph{Aligned patch grids.} To let the tools compare encoders patch by patch, we place the features of VGGT and DINOv3 on a shared $h \times w$ grid with $w = 37$. Each photo is resized to a width of 518\,px with its aspect ratio preserved, and a portrait photo is center-cropped to a square. VGGT-1B processes all $V$ views jointly on 14\,px patches, and we keep its aggregator patch tokens while discarding the camera token and the four register tokens, which yields $d_k = 2048$. DINOv3 ViT-L/16 encodes each photo separately after a resize that places its 16\,px patches on the same grid, with $d_k = 1024$. CLIP ViT-L/14 encodes the whole photo resized to 224\,px square on a coarser $16 \times 16$ grid with $d_k = 1024$, which the tools map onto the shared grid when comparing it with the other two encoders.

\paragraph{Direct input as feature images.}
\label{sec:method_phase1}
Direct input places the latent features $F^{(k)}$ next to the photos. The \textbf{global min--max image} draws the patches$\times$channels matrix as a single grayscale image, min--max normalized over the whole matrix. This rendering presents the features as an appearance, whereas their spatial information lies in the relations between patches, which motivates the tools in \Cref{sec:method_tools}.

\subsection{Basis-Invariant Latent Tools}
\label{sec:method_tools}
\label{sec:method_phase2}

\paragraph{Basis-invariant patch comparison.}
Since the channels of an external encoder carry no meaning that the VLM shares, we compare patches only within each encoder's feature space and rely solely on quantities that do not depend on the choice of channel basis. Inner products, Euclidean distances and feature norms are unchanged under any orthogonal change of basis $R \in \mathrm{O}(d_k)$, and so is a score that contrasts a patch with two reference regions,
\begin{equation}
    (R f(p))^\top (R \mu_A - R \mu_B) = f(p)^\top (\mu_A - \mu_B),
    \label{eq:rotation_invariance}
\end{equation}
where $f(p) \in \mathbb{R}^{d_k}$ is the feature of patch $p$, and $\mu_A$ and $\mu_B$ are the mean features of two regions. All tools are built from such quantities, with one exception. CLIP's \texttt{text\_map} compares patches with a text phrase, which works because CLIP embeds images and text in a shared space.

\paragraph{Cross-encoder consensus.}
The three encoders are trained with different objectives, so their errors need not coincide. The cross-encoder tools therefore measure with every applicable encoder and report how many of them agree, such as the direction vote of \texttt{motion} or the supported cells of \texttt{locate}. The analyst later uses this consensus to rate each finding in its report as high, medium or low (\Cref{sec:method_analyst}).

\begin{figure*}[t]
    \centering
    \includegraphics[width=0.98\textwidth]{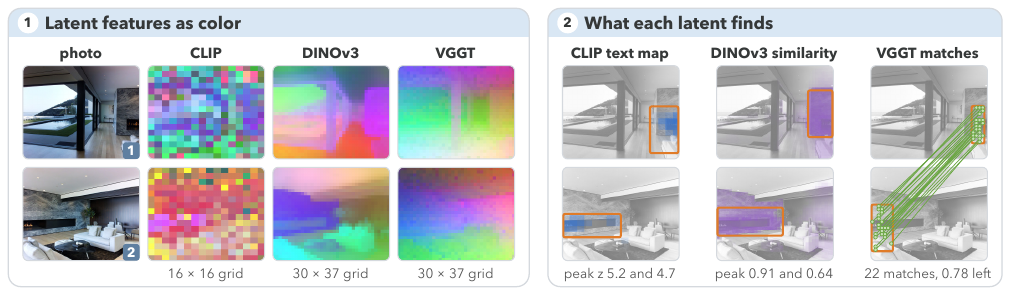}
    \caption{\mascot\ \textbf{What each latent feature picks out.} MMSI-Bench question 439. \textbf{(1)} Each encoder's first three principal components shown as color. \textbf{(2)} CLIP's text map finds the fireplace in both photos, DINOv3 marks patches similar to the region found in photo 1, and VGGT matches 22 of its patches into photo 2. Maps are drawn over the photos for the reader, and the analyst never sees a photo.}
    \label{fig:latent_roles}
\end{figure*}
\paragraph{Object localization by encoder consensus.}
Localization proceeds in three stages. CLIP proposes a region for each object name, DINOv3 and VGGT verify it, and their consensus sets a confidence tier. CLIP's proposal comes from \texttt{text\_map}, which scores each patch against the prompt ``a photo of $\langle$\textit{object}$\rangle$'' and subtracts its mean score for five generic phrases. Because last-layer CLIP patch tokens are known to score a named object below its surroundings~\cite{li2023clipsurgery}, the map is smoothed and negated. \texttt{locate} then z-scores the map across all photos and selects the connected component of the top 8\% of cells that contains the highest peak. \texttt{support} uses DINOv3 and VGGT to check which cells belong to the region, and cells supported by at least two encoders define its extent. Encoder consensus, region area, CLIP peak score and prominence set a confidence tier of high, medium, low or none, and \texttt{locate\_all} applies this procedure to every named object. \Cref{fig:latent_roles} shows what \texttt{text\_map}, DINOv3 similarity and VGGT matches find.

\paragraph{Cross-view motion and tracking.}
Two tools match patches across views. Given two photos, \texttt{motion} matches patches with the latent features of each encoder and reports how far the matched content shifts, whether it grows or shrinks, and the direction on which the encoders agree, such as ``left 3/3'' when all three agree. These quantities describe how the image changes rather than the camera pose. \texttt{track} instead follows a located object into the other photos using VGGT and DINOv3 patch matches, and CLIP checks the matched region wherever both encoders agree.

\paragraph{Within-view relations and layout.}
Within a single photo, \texttt{relate} computes the relative offset, size ratio and contact of two located objects. \texttt{layout} applies $k$-means with $k{=}5$ to VGGT and DINOv3 features, retains the segments that the two encoders share, and compares all three encoders with linear CKA~\cite{kornblith2019cka}. Other tools in \Cref{tab:tools} provide basic feature and geometric operations.

\begin{table}[t]
\centering
\caption{\textbf{Sandbox tools.} In \method\ the analyst picks tools and arguments. \emph{Runs on} lists the fixed arguments of the Fixed-Sequence ablation. Tools that compare photos need several photos. C, D and V are CLIP, DINOv3 and VGGT.}
\label{tab:tools}
\scriptsize
\setlength{\tabcolsep}{3pt}
\setlength{\aboverulesep}{0pt}\setlength{\belowrulesep}{0pt}\renewcommand{\arraystretch}{1.12}
\begin{tabularx}{\columnwidth}{@{}L l>{\raggedright\arraybackslash}p{0.27\columnwidth}l@{}}
\toprule
Tool & Features & Output & Runs on \\
\midrule
\texttt{motion} & C, D, V & shift, zoom, vote & $\le$7 photo pairs \\
\texttt{locate\_all}, \texttt{locate} & C, D, V & region and tier & each object \\
\texttt{support} & D, V & cells grouped with a seed & inside \texttt{locate} \\
\texttt{relate} & none & offset, size ratio, contact & located pairs \\
\texttt{track} & D, V & landing, CLIP check & $\le$3 other photos \\
\texttt{layout} & C, D, V & shared segments, CKA & $\le$3 photos \\
\midrule
\texttt{text\_map}, \texttt{top\_region}, \texttt{describe}, \texttt{region\_stats} & C & object map, blob & first object \\
\texttt{sim\_to}, \texttt{grow}, \texttt{pca}, \texttt{kmeans} & D, V & similarity, area, variance, segments & each encoder \\
\texttt{match} & C, D, V & patch matches & photos 1 and 2 \\
\texttt{grid}, \texttt{box}, \texttt{bbox}, \texttt{iou}, \texttt{corr}, \texttt{to\_vggt}, \texttt{to\_clip} & none & geometry, overlap, correlation & once \\
\rowcolor{tabgray}
\texttt{show} & any & map drawing & not available \\
\bottomrule
\end{tabularx}
\end{table}
\providecommand{\dup}[1]{\textcolor{ForestGreen}{+#1}}
\providecommand{\ddn}[1]{\textcolor{BrickRed}{$-$#1}}
\definecolor{besthl}{HTML}{D6E9F8}
\begin{table*}[t]
\centering
\caption{\mascot\ \textbf{Main results of five VLMs on three spatial benchmarks.} Accuracy in \% with the Correct latent features. \textcolor{gray}{Direct} gives the features directly and Agent parses them in the Agent-Directed mode, which with Multi-Latent features is \method. Shaded rows are photo only and $\Delta$ is Agent minus photo only. Blue cells mark each VLM's largest gain per benchmark.}
\label{tab:main_factorial}
\footnotesize
\setlength{\tabcolsep}{2.4pt}
\setlength{\aboverulesep}{0pt}\setlength{\belowrulesep}{0pt}\renewcommand{\arraystretch}{1.12}
\begin{tabularx}{\textwidth}{@{}c|cl *{12}{Y}@{}}
\toprule
& & & \multicolumn{3}{c}{MMSI-Bench} & \multicolumn{3}{c}{MV-RoboBench} & \multicolumn{3}{c}{SPAR-Bench} & \multicolumn{3}{c}{Avg.} \\
\cmidrule(lr){4-6}\cmidrule(lr){7-9}\cmidrule(lr){10-12}\cmidrule(l){13-15}
& VLM & Input & \textcolor{gray}{Direct} & Agent & $\Delta$ & \textcolor{gray}{Direct} & Agent & $\Delta$ & \textcolor{gray}{Direct} & Agent & $\Delta$ & \textcolor{gray}{Direct} & Agent & $\Delta$ \\
\midrule
\rowcolor{tabgray}
\cellcolor{white} & \cellcolor{white} & Photo only & \multicolumn{3}{c}{34.4} & \multicolumn{3}{c}{27.9} & \multicolumn{3}{c}{57.4} & \multicolumn{3}{c}{39.9} \\
 &  & + CLIP & \textcolor{gray}{33.8} & 35.6 & \dup{1.2} & \textcolor{gray}{28.1} & 26.7 & \ddn{1.2} & \textcolor{gray}{59.3} & \cellcolor{besthl}58.7 & \cellcolor{besthl}\dup{1.3} & \textcolor{gray}{40.4} & 40.3 & \dup{0.4} \\
 &  & + DINOv3 & \textcolor{gray}{33.1} & 36.3 & \dup{1.9} & \textcolor{gray}{29.2} & \cellcolor{besthl}33.1 & \cellcolor{besthl}\dup{5.2} & \textcolor{gray}{58.7} & 55.1 & \ddn{2.3} & \textcolor{gray}{40.3} & 41.5 & \dup{1.6} \\
 &  & + VGGT & \textcolor{gray}{33.1} & 38.5 & \dup{4.1} & \textcolor{gray}{28.5} & 28.9 & \dup{1.0} & \textcolor{gray}{59.8} & 57.6 & \dup{0.2} & \textcolor{gray}{40.5} & 41.7 & \dup{1.8} \\
\dashsep{3-15}
 & \multirow{-5}{*}{\vlabel{Qwen3.8-27B}} & \textbf{All} (Multi-Latent) & \textcolor{gray}{33.0} & \cellcolor{besthl}39.5 & \cellcolor{besthl}\dup{5.1} & \textcolor{gray}{28.9} & 29.7 & \dup{1.8} & \textcolor{gray}{58.0} & 57.6 & \dup{0.2} & \textcolor{gray}{40.0} & \cellcolor{besthl}42.3 & \cellcolor{besthl}\dup{2.4} \\
\cmidrule{2-15}
\rowcolor{tabgray}
\cellcolor{white} & \cellcolor{white} & Photo only & \multicolumn{3}{c}{30.0} & \multicolumn{3}{c}{37.7} & \multicolumn{3}{c}{57.6} & \multicolumn{3}{c}{41.8} \\
 &  & + CLIP & \textcolor{gray}{29.8} & 29.4 & \ddn{0.6} & \textcolor{gray}{34.7} & \cellcolor{besthl}38.8 & \cellcolor{besthl}\dup{1.1} & \textcolor{gray}{57.3} & 54.0 & \ddn{3.6} & \textcolor{gray}{40.6} & 40.7 & \ddn{1.1} \\
 &  & + DINOv3 & \textcolor{gray}{30.2} & \cellcolor{besthl}33.9 & \cellcolor{besthl}\dup{3.9} & \textcolor{gray}{35.0} & 36.0 & \ddn{1.7} & \textcolor{gray}{55.8} & 55.1 & \ddn{2.5} & \textcolor{gray}{40.3} & 41.7 & \ddn{0.1} \\
 &  & + VGGT & \textcolor{gray}{30.9} & \cellcolor{besthl}33.9 & \cellcolor{besthl}\dup{3.9} & \textcolor{gray}{34.4} & 34.8 & \ddn{2.9} & \textcolor{gray}{57.5} & \cellcolor{besthl}58.5 & \cellcolor{besthl}\dup{0.9} & \textcolor{gray}{40.9} & 42.4 & \dup{0.6} \\
\dashsep{3-15}
 & \multirow{-5}{*}{\vlabel{Gemma 4}} & \textbf{All} (Multi-Latent) & \textcolor{gray}{29.3} & 32.6 & \dup{2.6} & \textcolor{gray}{34.0} & 37.8 & \dup{0.1} & \textcolor{gray}{57.1} & 58.4 & \dup{0.8} & \textcolor{gray}{40.1} & \cellcolor{besthl}42.9 & \cellcolor{besthl}\dup{1.1} \\
\cmidrule{2-15}
\rowcolor{tabgray}
\cellcolor{white} & \cellcolor{white} & Photo only & \multicolumn{3}{c}{28.7} & \multicolumn{3}{c}{22.8} & \multicolumn{3}{c}{48.7} & \multicolumn{3}{c}{33.4} \\
 &  & + CLIP & \textcolor{gray}{30.7} & 27.7 & \ddn{1.0} & \textcolor{gray}{22.9} & 20.8 & \ddn{2.0} & \textcolor{gray}{48.2} & \cellcolor{besthl}51.3 & \cellcolor{besthl}\dup{2.6} & \textcolor{gray}{33.9} & 33.3 & \ddn{0.1} \\
 &  & + DINOv3 & \textcolor{gray}{29.8} & 27.4 & \ddn{1.3} & \textcolor{gray}{21.7} & 19.4 & \ddn{3.4} & \textcolor{gray}{47.6} & 49.6 & \dup{0.9} & \textcolor{gray}{33.0} & 32.1 & \ddn{1.3} \\
 &  & + VGGT & \textcolor{gray}{29.8} & 27.8 & \ddn{0.9} & \textcolor{gray}{23.2} & 20.6 & \ddn{2.2} & \textcolor{gray}{48.2} & 48.2 & \ddn{0.5} & \textcolor{gray}{33.7} & 32.2 & \ddn{1.2} \\
\dashsep{3-15}
\multirow{-15}{*}{\rotatebox[origin=c]{90}{\scriptsize\textit{Open-source}}} & \multirow{-5}{*}{\vlabel{InternVL}} & \textbf{All} (Multi-Latent) & \textcolor{gray}{28.1} & \cellcolor{besthl}28.9 & \cellcolor{besthl}\dup{0.2} & \textcolor{gray}{20.9} & 22.5 & \ddn{0.3} & \textcolor{gray}{46.4} & 51.1 & \dup{2.4} & \textcolor{gray}{31.8} & \cellcolor{besthl}34.2 & \cellcolor{besthl}\dup{0.8} \\
\midrule
\rowcolor{tabgray}
\cellcolor{white} & \cellcolor{white} & Photo only & \multicolumn{3}{c}{38.4} & \multicolumn{3}{c}{34.7} & \multicolumn{3}{c}{54.0} & \multicolumn{3}{c}{42.4} \\
 &  & + CLIP & \textcolor{gray}{35.2} & 39.2 & \dup{0.8} & \textcolor{gray}{33.0} & 35.0 & \dup{0.3} & \textcolor{gray}{50.0} & 52.7 & \ddn{1.3} & \textcolor{gray}{39.4} & 42.3 & \ddn{0.1} \\
 &  & + DINOv3 & \textcolor{gray}{34.1} & 39.2 & \dup{0.8} & \textcolor{gray}{30.9} & \cellcolor{besthl}38.8 & \cellcolor{besthl}\dup{4.1} & \textcolor{gray}{49.5} & 50.5 & \ddn{3.5} & \textcolor{gray}{38.2} & 42.8 & \dup{0.4} \\
 &  & + VGGT & \textcolor{gray}{33.1} & 37.8 & \ddn{0.6} & \textcolor{gray}{30.9} & 37.7 & \dup{3.0} & \textcolor{gray}{48.7} & 53.8 & \ddn{0.2} & \textcolor{gray}{37.6} & \cellcolor{besthl}43.1 & \cellcolor{besthl}\dup{0.7} \\
\dashsep{3-15}
 & \multirow{-5}{*}{\vlabel{GPT-6 Luna}} & \textbf{All} (Multi-Latent) & \textcolor{gray}{35.8} & \cellcolor{besthl}40.6 & \cellcolor{besthl}\dup{2.2} & \textcolor{gray}{30.9} & 35.3 & \dup{0.6} & \textcolor{gray}{51.6} & 53.5 & \ddn{0.5} & \textcolor{gray}{39.4} & \cellcolor{besthl}43.1 & \cellcolor{besthl}\dup{0.7} \\
\cmidrule{2-15}
\rowcolor{tabgray}
\cellcolor{white} & \cellcolor{white} & Photo only & \multicolumn{3}{c}{38.7} & \multicolumn{3}{c}{34.4} & \multicolumn{3}{c}{55.2} & \multicolumn{3}{c}{42.8} \\
 &  & + CLIP & \textcolor{gray}{38.7} & 38.5 & \ddn{0.2} & \textcolor{gray}{33.2} & 30.8 & \ddn{3.6} & \textcolor{gray}{56.4} & 55.5 & \dup{0.3} & \textcolor{gray}{42.8} & 41.6 & \ddn{1.2} \\
 &  & + DINOv3 & \textcolor{gray}{41.4} & 42.9 & \dup{4.2} & \textcolor{gray}{32.5} & 32.9 & \ddn{1.5} & \textcolor{gray}{53.3} & \cellcolor{besthl}59.5 & \cellcolor{besthl}\dup{4.3} & \textcolor{gray}{42.4} & \cellcolor{besthl}45.1 & \cellcolor{besthl}\dup{2.3} \\
 &  & + VGGT & \textcolor{gray}{38.9} & \cellcolor{besthl}43.1 & \cellcolor{besthl}\dup{4.4} & \textcolor{gray}{32.4} & \cellcolor{besthl}35.0 & \cellcolor{besthl}\dup{0.6} & \textcolor{gray}{55.5} & 55.8 & \dup{0.6} & \textcolor{gray}{42.3} & 44.6 & \dup{1.8} \\
\dashsep{3-15}
\multirow{-10}{*}{\rotatebox[origin=c]{90}{\scriptsize\textit{Closed-source}}} & \multirow{-5}{*}{\vlabel{Claude Haiku}} & \textbf{All} (Multi-Latent) & \textcolor{gray}{39.3} & 41.3 & \dup{2.6} & \textcolor{gray}{29.8} & \cellcolor{besthl}35.0 & \cellcolor{besthl}\dup{0.6} & \textcolor{gray}{51.6} & 54.9 & \ddn{0.3} & \textcolor{gray}{40.2} & 43.7 & \dup{0.9} \\
\bottomrule
\end{tabularx}
\end{table*}

\begin{table}[!t]
\centering
\caption{\textbf{Agent-Directed vs.\ Fixed-Sequence mode.} Qwen3.8-27B on MMSI-Bench with Multi-Latent features. Acc.\ is accuracy in \% and s/q is seconds of analysis per question on one GPU over the same 25 questions. Fixed-Sequence runs six predefined code blocks in order.}
\label{tab:ablations}
\footnotesize
\setlength{\tabcolsep}{4pt}
\setlength{\aboverulesep}{0pt}\setlength{\belowrulesep}{0pt}\renewcommand{\arraystretch}{1.12}
\begin{tabularx}{\columnwidth}{l YY}
\toprule
Analysis & Acc. & s/q \\
\midrule
\rowcolor{tabgray}
Photo only & 34.4 & 0 \\
\textbf{Agent-Directed}, ours & 39.5 & 99 \\
\dashsep{1-3}
Fixed-Sequence & 38.9 & 79 \\
\rowcolor{tabblue}
$\Delta$ Agent-Directed $-$ Fixed-Sequence & +0.6 & +20 \\
\bottomrule
\end{tabularx}
\end{table}

\paragraph{Tool calibration.}
Several design choices in the tools are specified manually rather than learned, including the negated CLIP map, the top 8\% of cells in \texttt{locate} and $k{=}5$ in \texttt{layout}. The thresholds that define the confidence tiers and decide whether an encoder counts as reliable in \texttt{motion} are chosen once on 150 calibration questions (100 from 3DSRBench and 50 from MMSI-Bench) outside the development sample $s_1$ (\Cref{sec:experiments}) and reused in all experiments. They are set so that most high-tier boxes cover the named object, as judged by a VLM that sees the photo, and so that Gaussian noise almost never reaches the high tier or passes the reliability check. No answers are used in this process, and although the tools avoid learned decoding heads, they still involve task-specific design.

\subsection{Photo-Blind Analyst and Answering}
\label{sec:method_analyst}

\paragraph{Inputs of the photo-blind analyst.}
The analyst receives the question, the answer options, the number of views, the image dimensions and the latent features of CLIP, DINOv3 and VGGT, which we refer to as Multi-Latent features. It never sees the photos directly: it queries the features with tools, although its interpretation may also draw on the question and options. On benchmarks that mark targets with colored boxes or points, a rule-based color detector supplies the color and position of each marker (\Cref{sec:exp_axes}).

\paragraph{Agent-Directed tool use.}
In the \textbf{Agent-Directed mode}, the analyst receives no predefined object list. It selects tools and their arguments from \Cref{tab:tools} and executes them in a sandbox over up to four steps, each of which can build on the printed outputs of earlier steps. A heuristic rule in the analyst's prompt reads the image shifts from \texttt{motion} as camera motion, so that content moving left indicates that the camera turned or moved right. Image shifts alone do not determine the 3D motion uniquely. The analyst cannot call \texttt{show} and never views a rendering of the features. It completes a single pass without a critic, as illustrated in \Cref{fig:pipeline}. This loop follows existing code-execution harnesses~\cite{lee2026metaharness,shao2026harnessr1,wei2026selfevolvingharness}, and differs in its tool library and restriction to latent features.

\paragraph{Report-based answering.}
After its final step, the analyst writes a report of up to six findings followed by a brief summary, where each finding is supported by numerical evidence and rated high, medium or low by cross-encoder consensus. The same VLM then answers the question in a fresh context from the photos, the question and this report, without access to the tools or the latent features. The answer therefore depends on the analysis only through the report.

\paragraph{Fixed-Sequence tool execution.}
To isolate the effect of the analyst's tool choices, the \textbf{Fixed-Sequence mode} in \Cref{tab:ablations} keeps the same tools and report format but removes the choice of tools from the analyst. A text-only call to the same VLM first extracts up to four object names from the question and options. Six code blocks then run in a predetermined order with the arguments listed in \Cref{tab:tools}, where \texttt{motion} uses up to seven photo pairs starting from the nearest and \texttt{track} runs at most eight times per question. The analyst then reads the outputs and writes the report.

\section{Experiments}
\label{sec:experiments}\label{sec:exp_axes}
\begin{table}[t]
\centering
\caption{\textbf{Controls.} Qwen3.8-27B on MMSI-Bench with \method\ and with the Multi-Latent features given directly. Mismatched features come from another MMSI-Bench question, and Random Noise is Gaussian with each view's per-channel mean and standard deviation.}
\label{tab:app_noise}
\footnotesize
\setlength{\tabcolsep}{3pt}
\setlength{\aboverulesep}{0pt}\setlength{\belowrulesep}{0pt}\renewcommand{\arraystretch}{1.12}
\begin{tabularx}{\columnwidth}{l YY}
\toprule
& \method & Direct input \\
\cmidrule(lr){2-2}\cmidrule(l){3-3}
Features & Acc. & Global \\
\midrule
\rowcolor{tabgray}
Photo only & 34.4 & 34.4 \\
\textbf{Correct}, ours & 39.5 & 33.0 \\
\dashsep{1-3}
Mismatched features & 33.3 & 34.6 \\
Random Noise & 36.6 & 34.7 \\
\rowcolor{tabblue}
Correct $-$ Mismatched & +6.2 & $-$1.6 \\
\bottomrule
\end{tabularx}
\end{table}

\label{sec:exp_setup}

\paragraph{Implementation Details.}
Each VLM is both the analyst and the answering model. The analyst runs one round of up to four tool steps within 600\,s per question. Open VLMs decode greedily with thinking off and at most 1,000 new tokens per analyst step, each on one 80\,GB GPU such as the H100. The VLMs are Qwen3.8-27B, Gemma~4 (gemma-4-31B-it), InternVL3-38B, GPT-6 Luna (gpt-6-luna) and Claude Haiku (claude-haiku-5-5); the two closed VLMs run through their APIs. Tool settings follow \Cref{tab:tools}, and the report holds up to six findings. \Cref{sec:app_settings,sec:app_prompts} list every threshold and the prompts.

\paragraph{Baseline Methods.}
\emph{Photo only} gives the VLM the photos, question and answer options. \emph{Direct input}, the Direct columns of \Cref{tab:main_factorial}, adds the global min--max image of one encoder's latent features, or of all three, next to the photos. In the rows + CLIP, + DINOv3 and + VGGT, the Agent columns come from an earlier single-encoder analyst, which is also photo-blind but has only the basic tools in the lower block of \Cref{tab:tools} and can view the maps it draws on the feature grid with \texttt{show}. In the row All, \method\ parses the Multi-Latent features. For Qwen3.8-27B on MMSI-Bench we also test whether a gain comes from the features' content or the extra input, by replacing the question's own features (\textbf{Correct}) with two controls. \textbf{Mismatched} uses the features of another question of the same benchmark with the same number of views, one fixed donor per question. \textbf{Random Noise} is Gaussian noise with each view's per-channel mean and standard deviation over patches, seeded per question.

\paragraph{Benchmarks.}
MMSI-Bench~\cite{yang2025mmsibench} has 1,000 multi-view questions, from camera motion to multi-step reasoning. From MV-RoboBench~\cite{feng2026mvrobobench} we take its 861 spatial questions, each with five options on synchronized views of one robot workspace in AgiBot World or BridgeV2. They ask which box marks the same object in another view, which boxed object is closest to grasp, where boxed objects sit in a $5{\times}5{\times}5$ grid and which of five photos the wrist camera took with a given head view. SPAR-Bench-Tiny~\cite{zhang2025spar} has 550 multiple-choice questions in 11 tasks of 50 on ScanNet frames, 350 of them with two or three views. MV-RoboBench and SPAR-Bench mark targets with colored boxes or points drawn into the images. Since the analyst never sees the photos, a fixed color detector with no learned model gives it the color and position of each marker. Every VLM answers every question of the three sets.

\paragraph{Evaluation Metrics.}
We report accuracy and give differences in percentage points (pp). Paired exact sign tests compare two conditions on the same questions, and a difference is significant when $p{<}0.05$. All tests are on Qwen3.8-27B; see \Cref{tab:app_stats} and \Cref{sec:app_source}. Ablations change one part of the default, \method\ with Qwen3.8-27B on MMSI-Bench. During development we inspected answers only on a sample $s_1$ of 145 questions (100 from 3DSRBench and 45 from MMSI-Bench), so we also report MMSI-Bench results on the 955 questions outside it.

\subsection{Quantitative Results and Analysis}
\label{sec:exp_main}

\paragraph{\method\ improves all five VLMs on average.}
Across the three benchmarks in \Cref{tab:main_factorial}, \method\ raises mean accuracy over photo only for all five VLMs by $0.7$--$2.4$~pp, while giving the Multi-Latent features directly changes it by $-3.0$ to $+0.1$~pp. The gains are largest on MMSI-Bench. On SPAR-Bench, \method\ improves Qwen3.8-27B by only $0.2$~pp and lowers GPT-6 Luna and Claude Haiku by $0.5$ and $0.3$~pp. Multi-Latent features are not always the best choice, as Gemma 4 scores $1.0$~pp higher with CLIP alone on MV-RoboBench.

\paragraph{Mismatched features remove the gain.}
On MMSI-Bench, \method\ raises Qwen3.8-27B by $5.1$~pp over photo only, which scores $34.4\%$ in \Cref{tab:app_noise}. Mismatched features remove this gain and score $1.1$~pp below photo only, significantly below Correct. Random Noise keeps $2.2$~pp of the gain and is not significantly below Correct ($p{=}0.08$). Against a paired photo-only run the gain of Correct features is significant, and it stays at $5.8$~pp on the 955 MMSI-Bench questions outside the development sample $s_1$. In a separate run that varies the analyst's input (\Cref{sec:app_source}), reports from the question alone or the photos without tools do not differ significantly from photo only, whereas tools run on RGB patches beat it ($p{=}0.003$). All inputs gain on the 68 questions with a ``Cannot be determined'' or ``None of the above'' option, which inconclusive reports favor. Without them, \method\ beats photo only ($p{=}0.006$) and RGB patches ($p{=}0.033$) on camera questions, but not photo only on all 932 remaining questions ($p{=}0.48$). Answered counts differ, and the $p$-values are uncorrected.
With one encoder, the agent gains $4.1$~pp with VGGT, $1.9$~pp with DINOv3 and $1.2$~pp with CLIP in \Cref{fig:evidence}. VGGT alone is not significantly below all three encoders ($p{=}0.58$), but the two miss different questions, and at least one is right on $52.1\%$. The single-encoder runs use the earlier analyst and its tools, and so do not isolate the effect of adding encoders. \method\ is $17.5$~pp better on camera motion, while VGGT alone is $7.0$~pp better on multi-step reasoning.

\paragraph{The tested feature images do not help.}
Given directly, the Correct Multi-Latent features leave Qwen3.8-27B $1.4$~pp below photo only on MMSI-Bench, and Mismatched features or Random Noise stay within $0.3$~pp of photo only. Each single encoder given directly also scores below photo only there. On the development sample $s_1$, Correct and Mismatched features given directly differ by $0.7$~pp for CLIP and for VGGT ($p{=}1.00$ for both). \Cref{fig:radar} compares direct input with photo only across VLMs.

\paragraph{Camera questions gain and multi-step questions lose.}
On camera motion in MMSI-Bench, \method\ raises Qwen3.8-27B by $33.8$~pp in \Cref{fig:radar_types}. Over the 334 paired camera questions it gains $19.2$~pp, and over the 195 paired multi-step questions it loses $9.7$~pp, both significantly. On MV-RoboBench, grid position rises by $6.8$~pp and grasp distance falls by $3.0$~pp. \Cref{tab:app_types} lists every question type, and \Cref{fig:examples} shows one success and one failure. We read 35 reports on multi-step questions that photo only gets right and \method\ misses. In 14 the report points to a wrong option, often because a weak CLIP hit places an absent object in the reference photo or the analyst turns image left and right into another viewer's frame. In 18 it misses the asked relation, yet the answer changes.

\subsection{Ablation Study}
\label{sec:exp_ablations}

\paragraph{A fixed tool sequence comes close.}
The Fixed-Sequence mode runs six predefined code blocks instead of letting the analyst choose tools. On MMSI-Bench it scores $0.6$~pp below Agent-Directed in \Cref{tab:ablations}, and the two modes do not differ significantly ($p{=}0.73$). With VGGT alone the two modes tie. Fixed-Sequence also beats photo only significantly, by $5.1$~pp on the 989 questions with paired answers. On one GPU and the same 25 questions, analysis takes $99$\,s per question for Agent-Directed and $79$\,s for Fixed-Sequence. Feature extraction takes about $0.5$\,s and answering $1.5$\,s per question, so analysis dominates.

\paragraph{The two modes differ by question type.}
Agent-Directed beats Fixed-Sequence by $17.5$~pp on camera motion but trails by $7.1$~pp on multi-step reasoning over the 195 paired questions. The Agent-Directed analyst calls \texttt{motion} on $85\%$ of MMSI-Bench questions and rarely uses most basic tools, as \Cref{tab:app_tool_use} shows. On camera-motion questions it calls \texttt{motion} every time and \texttt{locate\_all} on $5\%$, while on object–object questions it calls \texttt{locate\_all} on $93\%$ and \texttt{relate} on $38\%$.

\begin{table}[t]
\centering
\caption{\textbf{Analyst that also sees the photo.} Accuracy in \% of Qwen3.8-27B on the 1,000 MMSI-Bench questions. Both analysts read the CLIP, DINOv3 and VGGT features, and the same VLM answers in the same way.}
\label{tab:app_see_photo}
\footnotesize
\setlength{\tabcolsep}{3pt}
\setlength{\aboverulesep}{0pt}\setlength{\belowrulesep}{0pt}\renewcommand{\arraystretch}{1.12}
\begin{tabularx}{\columnwidth}{l YYY}
\toprule
Analyst input & All & Camera motion & Multi-step \\
Questions & 1,000 & 74 & 198 \\
\midrule
\rowcolor{tabgray}
Photo only, no analyst & 34.4 & 37.8 & 36.4 \\
\textbf{Latent features}, \method & 39.5 & 71.6 & 26.3 \\
Photo and latent features & 39.3 & 64.9 & 34.8 \\
\bottomrule
\end{tabularx}
\end{table}
\paragraph{Seeing the photo helps multi-step questions.}
An analyst that also sees the photo scores within $0.2$~pp of \method\ in \Cref{tab:app_see_photo}. It recovers multi-step reasoning with a significant gain of $8.5$~pp, but trails by $6.7$~pp on camera motion, not significantly ($p{=}0.23$). Different questions may need different evidence.

\section{Conclusion}
\label{sec:conclusion}

We propose \method{}, a training-free framework that enables frozen VLMs to exploit the latent features of pretrained vision encoders for spatial reasoning. A photo-blind analyst parses the features of CLIP, DINOv3 and VGGT with basis-invariant tools and distills the evidence into a text report, which the VLM reads alongside the photos. While the tested grayscale feature images yield no consistent benefit, this report improves the mean accuracy of all five VLMs across three benchmarks, with the largest gains on MMSI-Bench camera-motion questions, suggesting that encoder latents become useful once the relations between their patches are made explicit. The Mismatched and Random Noise controls cover only Qwen3.8-27B on MMSI-Bench, multi-step reasoning declines without access to the photos, and letting the analyst choose its tools has not yet outperformed a predetermined sequence. Future work could combine photo and latent evidence adaptively per question.

\section*{LLM Usage Statement}
Large language models are also the subject of this study, since every VLM we test and the analyst of \method\ are built on one. We used Claude (Anthropic), through Claude Code, to write and debug the experiment code, launch and monitor runs, compute statistics from the run outputs, sort failure cases, draw figures and edit the text, including related work. We used ChatGPT (OpenAI) to give feedback on drafts and to edit the text. The authors reviewed all code, results and text and take full responsibility for the paper.

{
    \small
    \bibliographystyle{ieeenat_fullname}
    \bibliography{main}

\begin{thebibliography}{53}
\providecommand{\natexlab}[1]{#1}
\providecommand{\url}[1]{\texttt{#1}}
\expandafter\ifx\csname urlstyle\endcsname\relax
  \providecommand{\doi}[1]{doi: #1}\else
  \providecommand{\doi}{doi: \begingroup \urlstyle{rm}\Url}\fi

\bibitem[Chen et~al.(2024)Chen, Xu, Kirmani, Ichter, Driess, Florence, Sadigh, Guibas, and Xia]{chen2024spatialvlm}
Boyuan Chen, Zhuo Xu, Sean Kirmani, Brian Ichter, Danny Driess, Pete Florence, Dorsa Sadigh, Leonidas Guibas, and Fei Xia.
\newblock {SpatialVLM}: Endowing vision-language models with spatial reasoning capabilities.
\newblock In \emph{Proceedings of the IEEE/CVF Conference on Computer Vision and Pattern Recognition (CVPR)}, pages 14455--14465, 2024.

\bibitem[Chen et~al.(2025)Chen, Chen, Chen, Pons-Moll, and Xiu]{chen2025feat2gs}
Yue Chen, Xingyu Chen, Anpei Chen, Gerard Pons-Moll, and Yuliang Xiu.
\newblock {Feat2GS}: Probing visual foundation models with {Gaussian} splatting.
\newblock In \emph{CVPR}, 2025.

\bibitem[Cheng et~al.(2024)Cheng, Yin, Fu, Guo, Yang, Kautz, Wang, and Liu]{cheng2024spatialrgpt}
An-Chieh Cheng, Hongxu Yin, Yang Fu, Qiushan Guo, Ruihan Yang, Jan Kautz, Xiaolong Wang, and Sifei Liu.
\newblock {SpatialRGPT}: Grounded spatial reasoning in vision language models.
\newblock In \emph{Advances in Neural Information Processing Systems (NeurIPS)}, 2024.

\bibitem[El~Banani et~al.(2024)El~Banani, Raj, Maninis, Kar, Li, Rubinstein, Sun, Guibas, Johnson, and Jampani]{elbanani2024probe3d}
Mohamed El~Banani, Amit Raj, Kevis-Kokitsi Maninis, Abhishek Kar, Yuanzhen Li, Michael Rubinstein, Deqing Sun, Leonidas Guibas, Justin Johnson, and Varun Jampani.
\newblock Probing the {3D} awareness of visual foundation models.
\newblock In \emph{CVPR}, 2024.

\bibitem[Fan et~al.(2025)Fan, Zhang, Li, Zhang, Chen, Hu, Wang, Qu, Wang, Yan, et~al.]{fan2025vlm3r}
Zhiwen Fan, Jian Zhang, Renjie Li, Junge Zhang, Runjin Chen, Hezhen Hu, Kevin Wang, Huaizhi Qu, Dilin Wang, Zhicheng Yan, et~al.
\newblock {VLM-3R}: Vision-language models augmented with instruction-aligned {3D} reconstruction.
\newblock arXiv preprint arXiv:2505.20279, 2025.

\bibitem[Feng et~al.(2026)Feng, Kang, Wang, Du, Yan, Shi, Yuan, Liang, Deng, Li, Yang, An, Zheng, Wang, Chen, Xu, Liang, Yang, and Guo]{feng2026mvrobobench}
Zhiyuan Feng, Zhaolu Kang, Qijie Wang, Zhiying Du, Jiongrui Yan, Shubin Shi, Chengbo Yuan, Huizhi Liang, Yu Deng, Qixiu Li, Rushuai Yang, Arctanx An, Leqi Zheng, Weijie Wang, Shawn Chen, Sicheng Xu, Yaobo Liang, Jiaolong Yang, and Baining Guo.
\newblock Seeing across views: Benchmarking spatial reasoning of vision-language models in robotic scenes.
\newblock In \emph{International Conference on Learning Representations (ICLR)}, 2026.

\bibitem[Fu et~al.(2024)Fu, Hu, Li, Feng, Wang, Lin, Roth, Smith, Ma, and Krishna]{fu2024blink}
Xingyu Fu, Yushi Hu, Bangzheng Li, Yu Feng, Haoyu Wang, Xudong Lin, Dan Roth, Noah~A. Smith, Wei-Chiu Ma, and Ranjay Krishna.
\newblock {BLINK}: Multimodal large language models can see but not perceive.
\newblock In \emph{ECCV}, 2024.

\bibitem[Gupta and Kembhavi(2023)]{gupta2023visprog}
Tanmay Gupta and Aniruddha Kembhavi.
\newblock Visual programming: Compositional visual reasoning without training.
\newblock In \emph{Proceedings of the IEEE/CVF Conference on Computer Vision and Pattern Recognition (CVPR)}, pages 14953--14962, 2023.

\bibitem[Hao et~al.(2025)Hao, Sukhbaatar, Su, Li, Hu, Weston, and Tian]{hao2024coconut}
Shibo Hao, Sainbayar Sukhbaatar, DiJia Su, Xian Li, Zhiting Hu, Jason Weston, and Yuandong Tian.
\newblock Training large language models to reason in a continuous latent space.
\newblock In \emph{Conference on Language Modeling (COLM)}, 2025.

\bibitem[Hu et~al.(2024)Hu, Shi, Fu, Roth, Ostendorf, Zettlemoyer, Smith, and Krishna]{hu2024visualsketchpad}
Yushi Hu, Weijia Shi, Xingyu Fu, Dan Roth, Mari Ostendorf, Luke Zettlemoyer, Noah~A. Smith, and Ranjay Krishna.
\newblock Visual sketchpad: Sketching as a visual chain of thought for multimodal language models.
\newblock In \emph{Advances in Neural Information Processing Systems (NeurIPS)}, 2024.

\bibitem[Janjua et~al.(2026)Janjua, Silva, Niu, and Rashidi]{janjua2026perceptionprograms}
Muhammad~Kamran Janjua, Hugo Silva, Di Niu, and Bahador Rashidi.
\newblock Don't show pixels, show cues: Unlocking visual tool reasoning in language models via perception programs.
\newblock In \emph{Proceedings of the IEEE/CVF Conference on Computer Vision and Pattern Recognition (CVPR)}, 2026.

\bibitem[Jiang et~al.(2026)Jiang, Chu, Tian, Zhang, et~al.]{jiang2026harnessevolve}
Wen Jiang, Mingmin Chu, Yimeng Tian, Qianxin Zhang, et~al.
\newblock {HarnessEvolve}: Learning from reference trajectories for reliable agent self-evolution.
\newblock arXiv preprint arXiv:2609.00829, 2026.

\bibitem[Kamath et~al.(2023)Kamath, Hessel, and Chang]{kamath2023whatsup}
Amita Kamath, Jack Hessel, and Kai-Wei Chang.
\newblock What's ``up'' with vision-language models? {I}nvestigating their struggle with spatial reasoning.
\newblock In \emph{EMNLP}, 2023.

\bibitem[Karten et~al.(2026)Karten, Zhang, Upaa, Feng, et~al.]{karten2026continualharness}
Seth Karten, Joel Zhang, Tersoo Upaa, Ruirong Feng, et~al.
\newblock Continual harness: Online adaptation for self-improving foundation agents.
\newblock arXiv preprint arXiv:2605.09998, 2026.

\bibitem[Kornblith et~al.(2019)Kornblith, Norouzi, Lee, and Hinton]{kornblith2019cka}
Simon Kornblith, Mohammad Norouzi, Honglak Lee, and Geoffrey Hinton.
\newblock Similarity of neural network representations revisited.
\newblock In \emph{Proceedings of the International Conference on Machine Learning (ICML)}, pages 3519--3529, 2019.

\bibitem[Lee et~al.(2026)Lee, Nair, Zhang, Lee, et~al.]{lee2026metaharness}
Yoonho Lee, Roshen Nair, Qizheng Zhang, Kangwook Lee, et~al.
\newblock {Meta-Harness}: End-to-end optimization of model harnesses.
\newblock arXiv preprint arXiv:2603.28052, 2026.

\bibitem[Li et~al.(2026)Li, Li, Huang, Kong, Zhou, Lu, Gong, Jia, Liang, and Huang]{li2026georeader}
Rong Li, Shijie Li, Jiangyong Huang, Lingdong Kong, Jiaming Zhou, Dongyue Lu, Zeying Gong, Baoxiong Jia, Junwei Liang, and Siyuan Huang.
\newblock {GeoReader}: Learning latent geometry for spatial reasoning in vision-language models.
\newblock arXiv preprint arXiv:2610.18694, 2026.

\bibitem[Li et~al.(2023)Li, Wang, Duan, and Li]{li2023clipsurgery}
Yi Li, Hualiang Wang, Yiqun Duan, and Xiaomeng Li.
\newblock {CLIP} surgery for better explainability with enhancement in open-vocabulary tasks.
\newblock arXiv preprint arXiv:2304.05653, 2023.

\bibitem[Lin et~al.(2026)Lin, Liu, Pan, Lin, et~al.]{lin2026ahe}
Jiahang Lin, Shichun Liu, Chengjun Pan, Lizhi Lin, et~al.
\newblock Agentic harness engineering: Observability-driven automatic evolution of coding-agent harnesses.
\newblock arXiv preprint arXiv:2604.25850, 2026.

\bibitem[Liu et~al.(2025)Liu, Dong, Wang, Ma, Tang, Tang, Rao, Ma, and Krishna]{liu2025coarse}
Benlin Liu, Yuhao Dong, Yiqin Wang, Zixian Ma, Yansong Tang, Luming Tang, Yongming Rao, Wei-Chiu Ma, and Ranjay Krishna.
\newblock Coarse correspondences boost spatial-temporal reasoning in multimodal language model.
\newblock In \emph{Proceedings of the IEEE/CVF Conference on Computer Vision and Pattern Recognition (CVPR)}, 2025.

\bibitem[Liu et~al.(2024)Liu, Li, Li, Li, Zhang, Shen, and Lee]{liu2024llavanext}
Haotian Liu, Chunyuan Li, Yuheng Li, Bo Li, Yuanhan Zhang, Sheng Shen, and Yong~Jae Lee.
\newblock {LLaVA-NeXT}: Improved reasoning, {OCR}, and world knowledge.
\newblock https://llava-vl.github.io/blog/2024-01-30-llava-next/, 2024.

\bibitem[Liu et~al.(2026{\natexlab{a}})Liu, Ye, Gao, Cao, et~al.]{liu2026solpi}
Haozhe Liu, Tian Ye, Sensen Gao, Qihang Cao, et~al.
\newblock {SoL-Pi}: Recursively scaling auto-research loops for efficient agent harness.
\newblock arXiv preprint arXiv:2609.20519, 2026{\natexlab{a}}.

\bibitem[Liu et~al.(2026{\natexlab{b}})Liu, Zhan, Zhou, Du, Liang, and Pfister]{liu2026sandboxvlm}
Yifan Liu, Fangneng Zhan, Kaichen Zhou, Yilun Du, Paul~Pu Liang, and Hanspeter Pfister.
\newblock Abstract {3D} perception for spatial intelligence in vision-language models.
\newblock In \emph{CVPR}, 2026{\natexlab{b}}.

\bibitem[Ma et~al.(2024)Ma, Lu, Cheng, Trigoni, and Markham]{ma2024spatialpin}
Chenyang Ma, Kai Lu, Ta-Ying Cheng, Niki Trigoni, and Andrew Markham.
\newblock {SpatialPIN}: Enhancing spatial reasoning capabilities of vision-language models through prompting and interacting {3D} priors.
\newblock In \emph{NeurIPS}, 2024.

\bibitem[Ma et~al.(2025)Ma, Chen, Zhang, Chou, Chen, de~Melo, and Yuille]{ma20243dsrbench}
Wufei Ma, Haoyu Chen, Guofeng Zhang, Yu-Cheng Chou, Jieneng Chen, Celso~M. de Melo, and Alan Yuille.
\newblock {3DSRBench}: A comprehensive {3D} spatial reasoning benchmark.
\newblock In \emph{Proceedings of the IEEE/CVF International Conference on Computer Vision (ICCV)}, 2025.

\bibitem[Man et~al.(2024)Man, Zheng, Bao, Hebert, Gui, and Wang]{man2024lexicon3d}
Yunze Man, Shuhong Zheng, Zhipeng Bao, Martial Hebert, Liang-Yan Gui, and Yu-Xiong Wang.
\newblock {Lexicon3D}: Probing visual foundation models for complex {3D} scene understanding.
\newblock In \emph{NeurIPS}, 2024.

\bibitem[Oquab et~al.(2024)Oquab, Darcet, Moutakanni, Vo, Szafraniec, Khalidov, Fernandez, Haziza, Massa, El-Nouby, et~al.]{oquab2023dinov2}
Maxime Oquab, Timoth{\'e}e Darcet, Th{\'e}o Moutakanni, Huy Vo, Marc Szafraniec, Vasil Khalidov, Pierre Fernandez, Daniel Haziza, Francisco Massa, Alaaeldin El-Nouby, et~al.
\newblock {DINOv2}: Learning robust visual features without supervision.
\newblock \emph{Transactions on Machine Learning Research (TMLR)}, 2024.

\bibitem[Pan et~al.(2026)Pan, Liu, Lin, Zeng, et~al.]{pan2026rho}
Wenbo Pan, Shujie Liu, Chin-Yew Lin, Jingying Zeng, et~al.
\newblock Evolving agents in the dark: Retrospective harness optimization via self-preference.
\newblock In \emph{Findings of the Association for Computational Linguistics: EMNLP}, 2026.

\bibitem[Radford et~al.(2021)Radford, Kim, Hallacy, Ramesh, Goh, Agarwal, Sastry, Askell, Mishkin, Clark, et~al.]{radford2021clip}
Alec Radford, Jong~Wook Kim, Chris Hallacy, Aditya Ramesh, Gabriel Goh, Sandhini Agarwal, Girish Sastry, Amanda Askell, Pamela Mishkin, Jack Clark, et~al.
\newblock Learning transferable visual models from natural language supervision.
\newblock In \emph{Proceedings of the International Conference on Machine Learning (ICML)}, pages 8748--8763, 2021.

\bibitem[Shaham et~al.(2024)Shaham, Schwettmann, Wang, Rajaram, Hernandez, Andreas, and Torralba]{shaham2024maia}
Tamar~Rott Shaham, Sarah Schwettmann, Franklin Wang, Achyuta Rajaram, Evan Hernandez, Jacob Andreas, and Antonio Torralba.
\newblock A multimodal automated interpretability agent.
\newblock In \emph{Proceedings of the International Conference on Machine Learning (ICML)}, 2024.

\bibitem[Shao et~al.(2026)Shao, Zhang, Li, Wang, et~al.]{shao2026harnessr1}
Shuai Shao, Kangning Zhang, Qingyao Li, Shijian Wang, et~al.
\newblock {Harness-R1}: Learning to edit executable runtime harnesses from agent failure trajectories.
\newblock arXiv preprint arXiv:2608.02276, 2026.

\bibitem[Sim{\'e}oni et~al.(2025)Sim{\'e}oni, Vo, Seitzer, Baldassarre, Oquab, Jose, Khalidov, Szafraniec, Yi, Ramamonjisoa, et~al.]{simeoni2025dinov3}
Oriane Sim{\'e}oni, Huy~V. Vo, Maximilian Seitzer, Federico Baldassarre, Maxime Oquab, Cijo Jose, Vasil Khalidov, Marc Szafraniec, Seungeun Yi, Micha{\"e}l Ramamonjisoa, et~al.
\newblock {DINOv3}.
\newblock arXiv preprint arXiv:2508.10104, 2025.

\bibitem[Sur{\'\i}s et~al.(2023)Sur{\'\i}s, Menon, and Vondrick]{suris2023vipergpt}
D{\'\i}dac Sur{\'\i}s, Sachit Menon, and Carl Vondrick.
\newblock {ViperGPT}: Visual inference via {Python} execution for reasoning.
\newblock In \emph{Proceedings of the IEEE/CVF International Conference on Computer Vision (ICCV)}, pages 11888--11898, 2023.

\bibitem[Tong et~al.(2024)Tong, Brown, Wu, Woo, Middepogu, Akula, Yang, Yang, Iyer, Pan, et~al.]{tong2024cambrian}
Shengbang Tong, Ellis Brown, Penghao Wu, Sanghyun Woo, Manoj Middepogu, Sai~Charitha Akula, Jihan Yang, Shusheng Yang, Adithya Iyer, Xichen Pan, et~al.
\newblock {Cambrian-1}: A fully open, vision-centric exploration of multimodal {LLMs}.
\newblock In \emph{Advances in Neural Information Processing Systems (NeurIPS)}, 2024.

\bibitem[Wang et~al.(2025)Wang, Chen, Karaev, Vedaldi, Rupprecht, and Novotny]{wang2025vggt}
Jianyuan Wang, Minghao Chen, Nikita Karaev, Andrea Vedaldi, Christian Rupprecht, and David Novotny.
\newblock {VGGT}: Visual geometry grounded transformer.
\newblock In \emph{Proceedings of the IEEE/CVF Conference on Computer Vision and Pattern Recognition (CVPR)}, 2025.

\bibitem[Wang et~al.(2026)Wang, Yin, Zhang, Zhang, Wang, Wang, Zhang, Chandrasegaran, Liu, Krishna, Xie, Wu, Fei-Fei, and Li]{wang2026mindcube}
Qineng Wang, Baiqiao Yin, Pingyue Zhang, Jianshu Zhang, Kangrui Wang, Zihan Wang, Jieyu Zhang, Keshigeyan Chandrasegaran, Han Liu, Ranjay Krishna, Saining Xie, Jiajun Wu, Li Fei-Fei, and Manling Li.
\newblock {MindCube}: Spatial mental modeling from limited views.
\newblock In \emph{International Conference on Learning Representations (ICLR)}, 2026.
\newblock arXiv:2506.21458.

\bibitem[Wei et~al.(2026)Wei, Guan, Zhu, Zhou, et~al.]{wei2026selfevolvingharness}
Chuyang Wei, Haoxiang Guan, Minghang Zhu, Kexin Zhou, et~al.
\newblock Self-evolving agent harness.
\newblock arXiv preprint arXiv:2610.18813, 2026.

\bibitem[Wong et~al.(2025)Wong, Qi, Wang, Hu, et~al.]{wong2025confucius}
Sherman Wong, Zhenting Qi, Zhaodong Wang, Nathan Hu, et~al.
\newblock Confucius code agent: Scalable agent scaffolding for real-world codebases.
\newblock arXiv preprint arXiv:2512.10398, 2025.

\bibitem[Wu et~al.(2025)Wu, Liu, Hung, and Duan]{wu2025spatialmllm}
Diankun Wu, Fangfu Liu, Yi-Hsin Hung, and Yueqi Duan.
\newblock {Spatial-MLLM}: Boosting {MLLM} capabilities in visual-based spatial intelligence.
\newblock arXiv preprint arXiv:2505.23747, 2025.

\bibitem[Xiao et~al.(2026)Xiao, Lu, Wang, Zhang, et~al.]{xiao2026reharness}
Yuxin Xiao, Zhiyun Lu, Jianyu Wang, Yangtian Zhang, et~al.
\newblock Recurrent harness for agentic continual learning.
\newblock arXiv preprint arXiv:2610.12668, 2026.

\bibitem[Xu et~al.(2026)Xu, Chen, CuiZhu, Mishra, et~al.]{xu2026harness2}
Ruiyao Xu, Yanfei Chen, Zhongying CuiZhu, Bhavana~Dalvi Mishra, et~al.
\newblock {Harness$^2$}: Recursive agent harnessing for an open world.
\newblock arXiv preprint arXiv:2610.14224, 2026.

\bibitem[Yang et~al.(2025{\natexlab{a}})Yang, Yang, Gupta, Han, Fei-Fei, and Xie]{yang2024thinking}
Jihan Yang, Shusheng Yang, Anjali~W. Gupta, Rilyn Han, Li Fei-Fei, and Saining Xie.
\newblock Thinking in space: How multimodal large language models see, remember, and recall spaces.
\newblock In \emph{Proceedings of the IEEE/CVF Conference on Computer Vision and Pattern Recognition (CVPR)}, 2025{\natexlab{a}}.

\bibitem[Yang et~al.(2025{\natexlab{b}})Yang, Xu, Xie, Yang, Li, Lin, Zhu, Chen, Duan, Yue, et~al.]{yang2025mmsibench}
Sihan Yang, Runsen Xu, Yiman Xie, Sizhe Yang, Mo Li, Jingli Lin, Chenming Zhu, Xiaochen Chen, Haodong Duan, Xiangyu Yue, et~al.
\newblock {MMSI-Bench}: A benchmark for multi-image spatial intelligence.
\newblock arXiv preprint arXiv:2505.23764, 2025{\natexlab{b}}.

\bibitem[Yao et~al.(2026)Yao, Tan, Liu, Li, et~al.]{yao2026harnessbench}
Yilun Yao, Xinyu Tan, Chao-Hsuan Liu, Yaoming Li, et~al.
\newblock {Harness-Bench}: Evaluating agent harnesses across models and realistic workflows.
\newblock arXiv preprint arXiv:2605.27922, 2026.

\bibitem[Ye et~al.(2026)Ye, Lu, Dong, Su, et~al.]{ye2026harnesszero}
Haoran Ye, Yuxing Lu, Haonan Dong, Zhaochen Su, et~al.
\newblock {Harness-Zero}: Harness distillation via agent-as-harness.
\newblock arXiv preprint arXiv:2609.24974, 2026.

\bibitem[Yeh et~al.(2026)Yeh, Qian, Wang, Ma, Tighe, and Xiao]{yeh2026beyond3dvqa}
Chun-Hsiao Yeh, Shengyi Qian, Manchen Wang, Yi Ma, Joseph Tighe, and Fanyi Xiao.
\newblock Beyond {3D} {VQAs}: Injecting {3D} spatial priors into vision-language models for enhanced geometric reasoning.
\newblock In \emph{Proceedings of the IEEE/CVF Conference on Computer Vision and Pattern Recognition (CVPR)}, 2026.

\bibitem[Zhang et~al.(2025{\natexlab{a}})Zhang, Wu, Li, Shang, Xia, Huang, Zhang, Dong, Zhang, Wang, Tan, and Wei]{zhang2025latentsketchpad}
Huanyu Zhang, Wenshan Wu, Chengzu Li, Ning Shang, Yan Xia, Yangyu Huang, Yifan Zhang, Li Dong, Zhang Zhang, Liang Wang, Tieniu Tan, and Furu Wei.
\newblock Latent sketchpad: Sketching visual thoughts to elicit multimodal reasoning in {MLLMs}.
\newblock arXiv preprint arXiv:2510.24514, 2025{\natexlab{a}}.

\bibitem[Zhang et~al.(2026)Zhang, Zhang, Li, Zhang, et~al.]{zhang2026selfharness}
Hangfan Zhang, Shao Zhang, Kangcong Li, Chen Zhang, et~al.
\newblock {Self-Harness}: Harnesses that improve themselves.
\newblock arXiv preprint arXiv:2606.09498, 2026.

\bibitem[Zhang et~al.(2025{\natexlab{b}})Zhang, Chen, Zhou, Xu, Huang, Mei, Chen, Yuan, Cai, Huang, Quan, Xu, and Zhang]{zhang2025spar}
Jiahui Zhang, Yurui Chen, Yanpeng Zhou, Yueming Xu, Ze Huang, Jilin Mei, Junhui Chen, Yu-Jie Yuan, Xinyue Cai, Guowei Huang, Xingyue Quan, Hang Xu, and Li Zhang.
\newblock From flatland to space: Teaching vision-language models to perceive and reason in {3D}.
\newblock In \emph{Advances in Neural Information Processing Systems (NeurIPS), Datasets and Benchmarks Track}, 2025{\natexlab{b}}.
\newblock arXiv:2503.22976.

\bibitem[Zhang et~al.(2022)Zhang, Zeng, Guo, and Li]{zhang2022depthclip}
Renrui Zhang, Ziyao Zeng, Ziyu Guo, and Yafeng Li.
\newblock Can language understand depth?
\newblock In \emph{ACM MM}, 2022.

\bibitem[Zheng et~al.(2025)Zheng, Huang, Li, and Wang]{zheng2025vgllm}
Duo Zheng, Shijia Huang, Yanyang Li, and Liwei Wang.
\newblock Learning from videos for {3D} world: Enhancing {MLLMs} with {3D} vision geometry priors.
\newblock arXiv preprint arXiv:2505.24625, 2025.

\bibitem[Zhou et~al.(2026)Zhou, Yang, Sun, Wu, et~al.]{zhou2026openvlharness}
Yu Zhou, Cheng-Fu Yang, Rui Sun, Di Wu, et~al.
\newblock {OpenVLHarness}: Real world visual agentic tool-use with multimodal memory.
\newblock arXiv preprint arXiv:2610.26860, 2026.

\bibitem[Zou et~al.(2026)Zou, Qiu, Li, Yang, Tieu, Lu, Shen, Tong, Choi, He, Zou, Wang, and Yang]{zou2025latentmas}
Jiaru Zou, Ruizhong Qiu, Gaotang Li, Xiyuan Yang, Katherine Tieu, Pan Lu, Ke Shen, Hanghang Tong, Yejin Choi, Jingrui He, James Zou, Mengdi Wang, and Ling Yang.
\newblock Latent collaboration in multi-agent systems.
\newblock In \emph{Proceedings of the International Conference on Machine Learning (ICML)}, 2026.

\end{thebibliography}
}

\clearpage
\appendix

\clearpage
\maketitlesupplementary
\makeatletter
\setlength{\@fptop}{0pt}\setlength{\@fpsep}{14pt}\setlength{\@fpbot}{0pt plus 1fil}\setlength{\@dblfptop}{0pt}\setlength{\@dblfpsep}{14pt}\setlength{\@dblfpbot}{0pt plus 1fil}
\makeatother
\lstdefinestyle{appprompt}{basicstyle=\ttfamily\scriptsize,breaklines=true,breakatwhitespace=false,columns=fullflexible,keepspaces=true,showstringspaces=false,frame=single,framesep=3pt,framerule=0.4pt,rulecolor=\color{gray!60},xleftmargin=3.4pt,xrightmargin=3.4pt,aboveskip=5pt,belowskip=5pt,breakindent=0pt,prebreak={},postbreak={}}
\section{Tool Settings}
\label{sec:app_settings}

\begin{table}[!h]
\centering
\caption{\textbf{Tool settings.} Fixed values of the tools in \Cref{tab:tools}, the same for every VLM and benchmark. A cell is one column of the shared grid, $1/37$ of the photo width, or one CLIP cell for CLIP's own matches. Peak and standout are z-scores of CLIP's text map.}
\label{tab:app_settings}
\footnotesize
\setlength{\tabcolsep}{3pt}
\setlength{\aboverulesep}{0pt}\setlength{\belowrulesep}{0pt}\renewcommand{\arraystretch}{1.12}
\begin{tabularx}{\columnwidth}{@{}l L@{}}
\toprule
Tool & Setting \\
\midrule
\texttt{text\_map} & subtracts the scores of ``background'', ``a wall'', ``the floor'', ``the sky'' and ``a surface'' \\
\texttt{locate} & CLIP blob from the top 8\% of cells and its core from the top 4\% \\
\texttt{support} & cells above the median plus 1.5 robust standard deviations of their similarity to the seed, each 1.4826 median absolute deviations \\
\texttt{locate} & an encoder keeps the blob when its support holds at least 0.3 of the core \\
High tier & both encoders keep the blob, VGGT--DINOv3 IoU at least 0.2, box area 0.02 to 0.40, peak at least 4.0, standout at least 1.0 \\
Medium tier & one encoder keeps the blob, box area 0.02 to 0.40, peak at least 3.0, standout at least 0.25 \\
\texttt{locate} & a high or medium blob drops to tier low when less than half of its CLIP cells lie inside the rows VGGT keeps \\
\texttt{locate\_all} & two names whose regions in one photo overlap with IoU at least 0.5 are one region, and the one with the lower peak drops to tier low \\
\texttt{motion} & an encoder is reliable with at least 30 mutual matches, or a quarter of its cells but never under 15, and with at least 0.20 of them fitting one shift, scale and rotation within 1.5 cells \\
\texttt{motion} & a direction needs a median shift of one cell, over the moving matches when at least 0.30 of them move, and 0.60 of these matches shifting the same way \\
\texttt{motion} & zoom shrinks below scale 0.93 and grows above 1.08 \\
\texttt{motion} & a high finding needs \texttt{pt\_agree} of at least 0.1 \\
\texttt{track} & found yes when each encoder has at least 5 mutual matches and its landings spread at most 1.5 times the region's own spread plus 0.03, and the landing points of the encoders lie within the larger of 0.1 and that spread \\
\texttt{layout} & $k$-means with $k{=}5$, and a VGGT segment and its best DINOv3 segment are shared when their IoU is at least 0.5 \\
\bottomrule
\end{tabularx}
\end{table}

\paragraph{The tier thresholds were fitted once and without any answer.}
The thresholds of the tiers of \texttt{locate}, the bar of \texttt{support} and the two bars of \texttt{motion} in \Cref{tab:app_settings} come from one calibration run on 3DSRBench and MMSI-Bench questions, made before \method\ ran on any full benchmark. The calibration reads no answer and no accuracy. For each question it names the objects as the analyst would, runs the tools on the Correct features and on Random Noise, and asks a VLM judge that sees the photo whether each box of \texttt{locate} lands on its object. The thresholds of \texttt{locate} admit the most objects whose high-tier boxes land on their object at least 65\% of the time and whose medium-tier boxes do so at least 50\% of the time, while at most 1\% of Noise objects reach the high tier and at most 5\% the high or medium tier. The inlier share of \texttt{motion} sits just above its 99th percentile of 0.18 under Noise, and the \texttt{pt\_agree} bar of 0.1 lies well above its 99th percentile of 0.02 under Noise.

\paragraph{At these settings boxes land on their objects and Noise yields none.}
On the calibration questions, 65\% of the high-tier boxes and 54\% of the medium-tier boxes land on their object, against 3\% for random boxes of the same size, and no Noise object reaches the high or medium tier.

\paragraph{No setting changed afterwards.}
All reported runs use these values, for all five VLMs, all three benchmarks and every question. MV-RoboBench and SPAR-Bench-Tiny entered the study only after the settings were frozen, so no value of \Cref{tab:app_settings} was fitted to either benchmark.

\section{Prompts}
\label{sec:app_prompts}

On MMSI-Bench, the analyst of \method\ receives the system prompt below word for word, for every question and VLM. Its first user message gives the number of photos, which it does not see, the grid sizes, the question and the options, and the sandbox holds the photo sizes as \texttt{photo\_size}. On MV-RoboBench and SPAR-Bench, this message also lists the color and position of each marker found by the color detector, followed by one of the two notes printed below. The printed output of each code step returns as the next user message under its execution id. Once four code steps or 600\,s are used, the analyst is asked at most twice to reply with findings only. The boxes it returns under \texttt{regions} are kept for inspection, as in \Cref{fig:examples_benches}, and the photos reach the answering VLM without them.

\begin{lstlisting}[style=appprompt]
You are the analyst. You do NOT see the photo(s). You get a spatial question and the raw latent features of the photo(s) from frozen vision models, loaded in a Python sandbox:
- F["clip"][v]: CLIP ViT-L/14 patch tokens, shape (16, 16, 1024): a 16x16 grid over the whole photo (squeezed to a square). CLIP links image regions to words.
- F["vggt"][v]: VGGT aggregator tokens, shape (h, 37, 2048). VGGT is a 3D reconstruction model: its features encode surfaces, depth and viewpoint, computed jointly for all photos of a scene.
- F["dinov3"][v]: DINOv3 ViT-L/16 patch tokens, shape (h, 37, 1024), on exactly the VGGT grid (same cells, same vggt_valid). DINOv3 is a self-supervised image model: features of objects, parts and surfaces, computed for each photo on its own (it never sees the other photos and has no notion of depth).
Nothing is decoded: there is no depth map, no 3D points and no camera pose. CLIP's grid is 16x16 over the whole photo; VGGT and DINOv3 share a grid of 14-pixel cells, 37 columns across the photo width (a tall photo is centre-cropped; vggt_valid[v] is False on padding rows). Positions are photo fractions: x from the left, y from the top, both 0-1. Printouts number the photos 1, 2, ...; in code and in "regions", view = photo number - 1.

Your job is NOT to answer. Measure evidence in the latents that bears on the question, check it across latents, and say how reliable it is. There is no critic and no second round; a separate model answers from the photo(s) and your findings.

Tools that compare latents (start with these):
- locate(phrase, v=None): CLIP's text map (already negated, so the named object scores highest) finds the object in the photo where it peaks (or in photo v); VGGT and DINOv3 each check on their own which cells belong with CLIP's blob. Prints photo, tier, box, area, centroid, peak, standout (peak minus the best value elsewhere), keep (share of the core of CLIP's blob each latent keeps), iou_vggt_dinov3 and peaks per photo. tier is high, medium, low or none. locate_all(phrases) runs it for each phrase and marks two names that land on one region (same_as).
- motion(va, vb): how the content moves from photo va to photo vb in each latent, from mutual patch matches: n, dx, dy (median shift of the moving matches; + = right / down), static (share not moving), inl (share that one shift + scale + rotation explains; noise stays under 0.2), scale (> 1: content grows), rot (degrees). reliable lists the latents with enough consistent matches (an unreliable one shows only n and inl); x, y and zoom give the direction of the reliable latents and how many agree ("left 3/3", "left 2/3 (clip right)", "left (only vggt)", "disagree", "mixed", "unreliable"); pt_agree is the share of cells VGGT and DINOv3 both match whose targets lie within one cell of each other (chance is about 0.01).
- track(region, vb): where a located region (a locate result) lands in photo vb in each geometry latent (n, at, spread); found yes / no / unclear; only with yes: shift, beyond_camera (shift minus the camera motion: the object's own movement, or parallax) and text_at vs text_peak (CLIP's map at the landing place vs its peak there).
- relate(a, b): region b relative to region a in one photo: dx, dy between centroids, size (box area ratio), iou, touch.
- layout(v): k-means segments VGGT and DINOv3 share, and cka: how alike each pair of latents' maps are (0 unrelated, 1 the same structure).
- support(model, seed, v): the cells a latent itself groups with a seed region (the seed is not forced in). iou(a, b, v): overlap of two masks of photo v (either grid).
Single-latent tools: text_map(phrase, v, pooled=False), top_region(map, frac), sim_to(model, mask, v, mask_view), grow(model, seed, v) (its result always contains the seed, so it never confirms a region), kmeans(model, v, k, mask), pca(model, v, k, mask), match(va, vb, model, mask_a), region_stats(map, mask, v), corr(map_a, map_b, mask), box(x0, y0, x1, y1, model, v), bbox(mask, v), to_vggt / to_clip(map_or_mask, v), describe(map), grid(model, v); photo_size, n_views and latents (the latents in F). numpy and scipy also work.

Rules for reading the numbers:
1. Rely on a location, a motion or a relation only when the latents agree, and set each finding's reliability from that agreement:
   - high: locate tier high; a motion direction that every reliable latent gives, with at least two of them ("left 3/3", "left 2/2") and pt_agree >= 0.1 when VGGT and DINOv3 are both reliable; track found yes.
   - medium: locate tier medium; a direction with a dissenting latent ("left 2/3 (...)") or from one latent ("only ..."); track "yes (only ...)".
   - low: anything else. "disagree", "mixed", "unreliable", tier low or none, and track no or unclear are not evidence: say the latents are inconclusive on that point. Never settle a disagreement by picking the latent you prefer.
2. Agreement shows that a region is one coherent thing in several latents, not that it is the named object; that rests on CLIP's peak and standout.
3. Camera motion: content moving right from photo a to photo b means the camera turned or moved left; content moving down means it tilted or moved up; zoom "grows" means it moved toward the scene. A shift under one cell (0.027) is no movement.
4. State a relation between two objects only when both have tier high or medium; left/right and above/below come from their centroids in one photo.
5. The latents carry no reliable signal for orientation: which way an object faces or points, or whether two objects are parallel or face the same way. Do not infer these; say they are not measured.
6. PCA scores, similarities and cluster ids have arbitrary offset and scale: compare orderings and differences relative to the spread, never ratios, metres or angles.
7. If nothing reaches medium reliability, say the latents are inconclusive. Every finding must come from the latent numbers (source "latent"); do not describe what the scene probably looks like.

Write Python: exactly one ```python block per turn, at most 4 code steps; variables persist between steps and each run gets an id (E1, E2, ...). Print short summaries (the tools above print compact lines), never whole arrays. Then give your FINDINGS.
In "regions", give the box of each object the question names whose tier is high, or medium when track found it in another photo (found yes); view = photo number - 1. Leave out every other object. The boxes are drawn on the photo(s) for the model that answers, so a wrong box misleads it.
When done, reply with a line starting with FINDINGS: followed by JSON (at most 6 findings, most decisive first, each at most 60 words):
{"findings": [{"text": "<finding with its numbers, the tool and the latents>", "source": "latent", "evidence": ["E2"], "reliability": "high|medium|low", "why": "<which latents agree or disagree>"}], "summary": "<at most 80 words on what the latents show that bears on the question; do not pick an option>", "regions": [{"label": "<object>", "view": 0, "box": [0.1, 0.2, 0.4, 0.6]}]}
\end{lstlisting}

\noindent The first user message has this form, with $h$ rows set by each photo's aspect ratio.

\begin{lstlisting}[style=appprompt]
2 photo(s) (not shown to you). Grids: CLIP (16, 16) per photo; VGGT and DINOv3 [(h, 37), (h, 37)].
Question: <question>
Options: A. <option>; B. <option>; C. <option>; D. <option>
\end{lstlisting}

\noindent On MV-RoboBench and SPAR-Bench, the marks follow the options in the form below, with the first note.

\begin{lstlisting}[style=appprompt]
Marks in the photo(s):
- Photo 1: red box, x 0.12-0.30, y 0.40-0.62
- Photo 2: blue box, x 0.55-0.71, y 0.20-0.44
These marks are drawn in the photo(s) and are what the question refers to. Their positions are given, not guessed, so a marked object counts as located: box(x0, y0, x1, y1, model, v) with these coordinates (v = photo number - 1; for a point, a box of +-0.04 around it) gives the latent cells of the marked object, usable with the other tools.
\end{lstlisting}

\noindent When the question names a reference view and no mark lies in that photo, the second note replaces the first.

\begin{lstlisting}[style=appprompt]
These marks are drawn in the photo(s) and are what the question refers to. Their positions are given, not guessed, so a marked object counts as located: box(x0, y0, x1, y1, model, v) with these coordinates (v = photo number - 1; for a point, a box of +-0.04 around it) gives the latent cells of the marked object as a boolean mask (print mask.sum(), not the mask), usable with the other tools. A mark's position holds only in the photo it is drawn in: when the question asks about a relation seen from another photo (for example the first photo as the observer's view), carry the marked cells there with track(mask, vb, va=v) or locate the object in that photo, and never read a position in one photo as a position in another.
\end{lstlisting}

\noindent When the step or time limit is reached, the analyst receives this message.

\begin{lstlisting}[style=appprompt]
Step or time limit reached. Do not write code. Reply only with FINDINGS: and the JSON, using the evidence you already have (say what is still uncertain).
\end{lstlisting}

\noindent The answering VLM receives the photos, the report under a fixed header and the question in the form of the photo-only runs, which receive the same message without the report. The report keeps the text and reliability of each finding and the summary, cut at 1,500 characters. Open VLMs continue after the prefix ``Answer:'', and the first option letter in the reply is the answer. Closed VLMs receive the same content as in their photo-only runs.

\begin{lstlisting}[style=appprompt]
<photo 1> <photo 2>
Notes from another model's analysis of the latent features (CLIP, VGGT and DINOv3) of the photo(s). They may contain errors:
- <finding> (reliability: high)
- <finding> (reliability: medium)
Summary: <summary>
<question>
A. <option>
B. <option>
C. <option>
D. <option>
Answer with the option's letter from the given choices directly.
\end{lstlisting}

\section{Tools in the Agent-Directed Mode}
\label{sec:app_tool_use}

\begin{table}[!h]
\centering
\caption{\textbf{Tools the Agent-Directed analyst calls.} Qwen3.8-27B with Multi-Latent features on the 1,000 MMSI-Bench questions. Questions in \% have at least one call of the tool, and calls per question count only those questions. Five tools, \texttt{grow}, \texttt{pca}, \texttt{corr}, \texttt{iou} and \texttt{to\_clip}, are not called.}
\label{tab:app_tool_use}
\footnotesize
\setlength{\aboverulesep}{0pt}\setlength{\belowrulesep}{0pt}\renewcommand{\arraystretch}{1.0}
\begin{tabularx}{\columnwidth}{l YYY}
\toprule
Tool & Questions in \% & Calls & Calls per question \\
\midrule
\texttt{motion} & 84.5 & 1,270 & 1.50 \\
\texttt{locate\_all} & 69.2 & 726 & 1.05 \\
\texttt{layout} & 59.5 & 1,122 & 1.89 \\
\texttt{locate} & 53.8 & 1,760 & 3.27 \\
\texttt{text\_map} & 25.7 & 725 & 2.82 \\
\texttt{track} & 25.0 & 444 & 1.78 \\
\texttt{top\_region} & 14.3 & 260 & 1.82 \\
\texttt{relate} & 12.6 & 388 & 3.08 \\
\texttt{box} & 7.9 & 210 & 2.66 \\
\texttt{match} & 5.2 & 93 & 1.79 \\
Eight other tools & under 2 each & 115 & -- \\
\bottomrule
\end{tabularx}
\end{table}

In the Agent-Directed mode on MMSI-Bench, the analyst runs $3.87$ tool steps per question on average. Most of its calls, counted in \Cref{tab:app_tool_use}, go to the top rows of \Cref{tab:tools}. \texttt{motion} appears in $85\%$ of the questions, including every camera-motion question. The analyst relies on the tools that compare encoders, as \texttt{motion}, \texttt{locate\_all}, \texttt{layout} and \texttt{locate} together make up $69\%$ of all calls.

\section{Direct Input across VLMs}
\label{sec:app_direct}

\begin{figure}[t]
    \centering
    \includegraphics[width=\columnwidth]{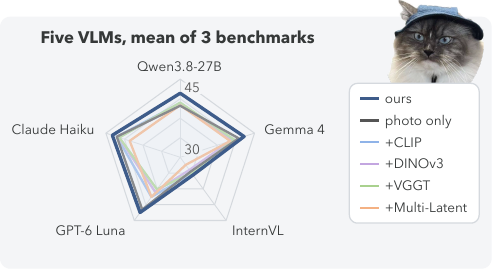}
    \caption{\mascot\ \textbf{Photo only against each encoder's features given directly.} Five VLMs on the mean of three benchmarks, with CLIP, DINOv3, VGGT or Multi-Latent features shown as an image next to the photo. The bold blue line is \method.}
    \label{fig:radar}
\end{figure}
\begin{figure*}[!t]
    \centering
    \includegraphics[width=0.98\textwidth]{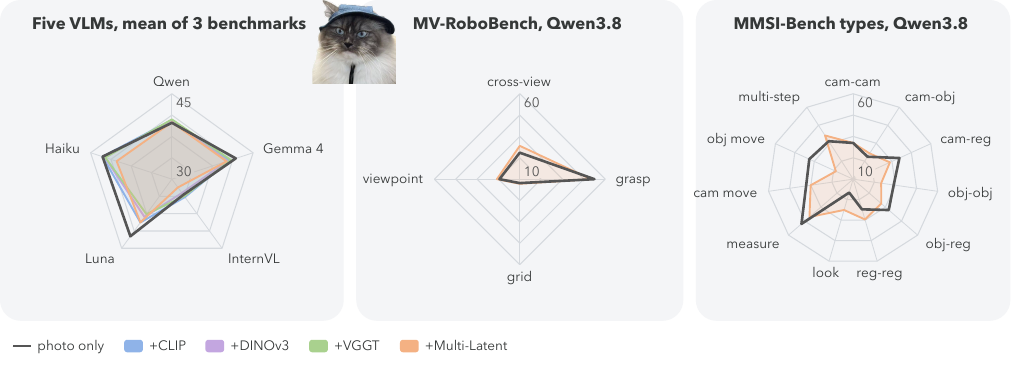}
    \caption{\mascot\ \textbf{Photo only and each encoder's features given directly, by question type.} Five VLMs on the mean of three benchmarks, as in \Cref{fig:radar}, and Qwen3.8-27B on each task of MV-RoboBench and each question type of MMSI-Bench. The radii start at 30\% in the left panel and at 10\% in the other two.}
    \label{fig:app_radar}
\end{figure*}

\Cref{fig:radar} compares photo only with the features of each encoder given directly, averaged over the three benchmarks, and \Cref{fig:app_radar} adds the question types of Qwen3.8-27B. Averaged over the three benchmarks, one encoder's features given directly change the accuracy of a VLM by $-4.8$ to $+0.6$~pp against photo only, and the Multi-Latent features lower four of the five VLMs by $1.6$ to $3.0$~pp and leave Qwen3.8-27B at $+0.1$~pp.

\section{Accuracy by Question Type}
\label{sec:app_types}
\Cref{tab:app_types} splits the accuracy of Qwen3.8-27B by question type on MV-RoboBench and MMSI-Bench. On MMSI-Bench, \method\ raises 7 of the 11 question types, five of them by more than 8~pp.

\begin{figure*}[t]
    \centering
    \includegraphics[width=0.98\textwidth]{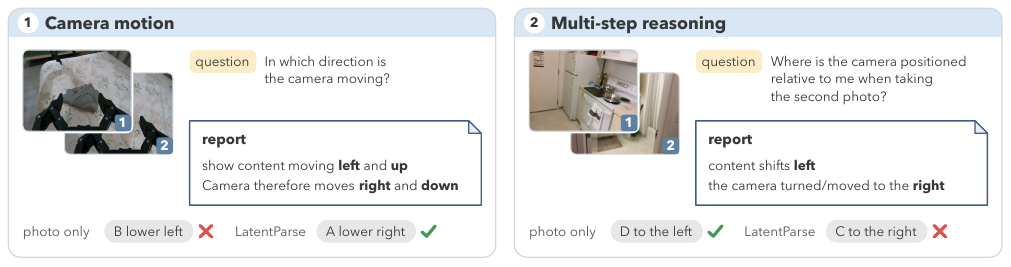}
    \caption{\mascot\ \textbf{Qualitative examples.} Two MMSI-Bench questions with Qwen3.8-27B. On camera motion, \method's report leads to the right answer where photo only fails. On a multi-step question, the analyst reads the leftward shift as the camera moving right, so \method\ answers wrong while photo only is right. Report lines are abridged excerpts.}
    \label{fig:examples}
\end{figure*}

\begin{table}[!h]
\centering
\caption{\textbf{Accuracy by question type.} Qwen3.8-27B on all questions of MV-RoboBench and MMSI-Bench. Photo only is the run of \Cref{tab:main_factorial}. \textbf{All} is \method\ with the Multi-Latent features, and $\Delta$ is its change from photo only. MVR is MV-RoboBench.}
\label{tab:app_types}
\footnotesize
\setlength{\tabcolsep}{2pt}
\setlength{\aboverulesep}{0pt}\setlength{\belowrulesep}{0pt}\renewcommand{\arraystretch}{1.0}
\begin{tabular*}{\columnwidth}{@{\extracolsep{\fill}}cl r c cc@{}}
\toprule
& Question type & $n$ & \shortstack{\rule{0pt}{8.5pt}Photo\\only} & \textbf{All} & $\Delta$ \\
\midrule
 & Cross-view matching & 200 & 25.5 & 25.5 & +0.0 \\
 & Grasp distance & 201 & 53.7 & 50.7 & $-$3.0 \\
 & Grid position & 204 & 12.3 & 19.1 & +6.8 \\
\multirow{-4}{*}{\vlabel{MVR}} & Viewpoint & 256 & 21.9 & 25.0 & +3.1 \\
\midrule
 & Positional: camera--camera & 93 & 31.2 & 37.6 & +6.4 \\
 & Positional: camera--object & 86 & 25.6 & 48.8 & +23.2 \\
 & Positional: camera--region & 83 & 39.8 & 48.2 & +8.4 \\
 & Positional: object--object & 94 & 34.0 & 30.9 & $-$3.1 \\
 & Positional: object--region & 85 & 37.6 & 37.6 & +0.0 \\
 & Positional: region--region & 81 & 28.4 & 37.0 & +8.6 \\
 & Attribute: appearance & 66 & 18.2 & 28.8 & +10.6 \\
 & Attribute: measurement & 64 & 50.0 & 48.4 & $-$1.6 \\
 & Motion: camera & 74 & 37.8 & 71.6 & +33.8 \\
 & Motion: object & 76 & 38.2 & 42.1 & +3.9 \\
\multirow{-11}{*}{\vlabel{MMSI-Bench}} & Multi-step reasoning & 198 & 36.4 & 26.3 & $-$10.1 \\
\bottomrule
\end{tabular*}
\end{table}

\section{Encoder Layers}
\label{sec:app_layers}

\begin{table}[t]
\centering
\caption{\textbf{Which encoder layer.} Accuracy in \% of Qwen3.8-27B with \method\ on 300 MMSI-Bench questions. Each column reads one encoder from another of its 24 blocks and keeps the other two at the last block. The last row concatenates blocks 5, 12, 18 and 24, which VGGT's DPT heads read.}
\label{tab:app_layers}
\footnotesize
\setlength{\tabcolsep}{3pt}
\setlength{\aboverulesep}{0pt}\setlength{\belowrulesep}{0pt}\renewcommand{\arraystretch}{1.12}
\begin{tabularx}{\columnwidth}{l YYY}
\toprule
Layer read & CLIP & DINOv3 & VGGT \\
\midrule
\rowcolor{tabgray}
Photo only & 29.7 & 29.7 & 29.7 \\
\textbf{Last block 24}, ours & 34.3 & 34.3 & 34.3 \\
\dashsep{1-4}
Early block 5 & 33.0 & 36.7 & 34.7 \\
Blocks 5, 12, 18 and 24 & 39.0 & 34.3 & 33.0 \\
\bottomrule
\end{tabularx}
\end{table}

\paragraph{\method\ is stable across encoder layers.}
We read one encoder at a time from another layer on 300 MMSI-Bench questions and keep the other two at their last block in \Cref{tab:app_layers}. Reading block 5, or blocks 5, 12, 18 and 24 together, changes accuracy by $-1.3$ to $+4.7$~pp against the last block, never significantly ($p{\ge}0.08$). With every layer choice, \method\ stays above photo only, at $33.0\%$ to $39.0\%$ against $29.7\%$.

\section{Paired Sign Tests}
\label{sec:app_stats}
\Cref{tab:app_stats} lists paired exact sign tests for the main experiments. Each test is two-sided and uses only the questions that exactly one of the two conditions answers correctly. W counts the questions that only the first condition answers correctly and L those that only the second does, written as 165W/103L in the sections below. The development sample $s_1$ holds 100 3DSRBench and 45 MMSI-Bench questions whose answers we inspected during development. Paired photo only is the answer setting of the agent runs that receives no notes, so it shares their GPUs and batches. The rows on 986 questions keep the questions on which its prompt is exactly the photo-only prompt, with no box drawn in any photo, as in \Cref{tab:app_more}. The Fixed-Sequence row keeps such questions of the Fixed-Sequence runs, 989 of them, and the other paired rows use all questions that both conditions answer.

\begin{table*}[t]
\centering
\caption{\textbf{Paired sign tests of the main text.} Qwen3.8-27B on MMSI-Bench, and on the development sample $s_1$ for direct input. Acc.\ is the accuracy in \% of the first and of the second condition on the $n$ questions that both answer.}
\label{tab:app_stats}
\footnotesize
\setlength{\tabcolsep}{5pt}
\setlength{\aboverulesep}{0pt}\setlength{\belowrulesep}{0pt}\renewcommand{\arraystretch}{1.12}
\begin{tabular*}{\textwidth}{@{\extracolsep{\fill}}l r cc rr r@{}}
\toprule
& & \multicolumn{2}{c}{Acc.} & & & \\
\cmidrule(lr){3-4}
First vs.\ second condition & $n$ & First & Second & W & L & $p$ \\
\midrule
\multicolumn{7}{@{}l}{\textit{\method\ and its controls}} \\
Correct vs.\ Mismatched features & 1,000 & 39.5 & 33.3 & 165 & 103 & 0.0002 \\
Correct vs.\ Random Noise & 1,000 & 39.5 & 36.6 & 141 & 112 & 0.08 \\
Correct vs.\ paired photo only & 1,000 & 39.5 & 33.7 & 184 & 126 & 0.001 \\
\quad outside $s_1$ & 955 & 39.6 & 33.8 & 176 & 121 & 0.002 \\
Correct vs.\ paired photo only, no box drawn & 986 & 39.5 & 33.7 & 181 & 124 & 0.001 \\
\quad camera questions & 334 & 50.6 & 31.4 & 99 & 35 & $<$10$^{-4}$ \\
\quad multi-step reasoning & 195 & 26.2 & 35.9 & 20 & 39 & 0.02 \\
VGGT alone vs.\ Multi-Latent features & 1,000 & 38.5 & 39.5 & 126 & 136 & 0.58 \\
\midrule
\multicolumn{7}{@{}l}{\textit{Direct input on $s_1$}} \\
Correct vs.\ Mismatched, CLIP & 145 & 55.9 & 55.2 & 2 & 1 & 1.00 \\
Correct vs.\ Mismatched, VGGT & 145 & 58.6 & 59.3 & 3 & 4 & 1.00 \\
\midrule
\multicolumn{7}{@{}l}{\textit{Ablations}} \\
Agent-Directed vs.\ Fixed-Sequence & 1,000 & 39.5 & 38.9 & 107 & 101 & 0.73 \\
\quad with VGGT alone & 1,000 & 38.5 & 38.5 & 105 & 105 & 1.00 \\
Fixed-Sequence vs.\ paired photo only & 989 & 38.8 & 33.7 & 166 & 115 & 0.003 \\
Photo and latent features vs.\ \method, multi-step reasoning & 198 & 34.8 & 26.3 & 32 & 15 & 0.02 \\
Photo and latent features vs.\ \method, camera motion & 74 & 64.9 & 71.6 & 3 & 8 & 0.23 \\
Block 5 vs.\ last block, CLIP & 300 & 33.0 & 34.3 & 25 & 29 & 0.68 \\
Block 5 vs.\ last block, DINOv3 & 300 & 36.7 & 34.3 & 35 & 28 & 0.45 \\
Block 5 vs.\ last block, VGGT & 300 & 34.7 & 34.3 & 32 & 31 & 1.00 \\
Blocks 5, 12, 18 and 24 vs.\ last block, CLIP & 300 & 39.0 & 34.3 & 35 & 21 & 0.08 \\
Blocks 5, 12, 18 and 24 vs.\ last block, DINOv3 & 300 & 34.3 & 34.3 & 27 & 27 & 1.00 \\
Blocks 5, 12, 18 and 24 vs.\ last block, VGGT & 300 & 33.0 & 34.3 & 30 & 34 & 0.71 \\
\bottomrule
\end{tabular*}
\end{table*}

\section{Source of the Gain on MMSI-Bench}
\label{sec:app_source}

\begin{table*}[t]
\centering
\caption{\textbf{What the analyst receives.} Accuracy in \% of Qwen3.8-27B on the camera questions of MMSI-Bench in a separate run, over the answered questions. The same VLM answers in the same way from the photos, the question and the report. Gray numbers are the $p$-values of paired exact sign tests against photo only.}
\label{tab:app_source}
\footnotesize
\setlength{\tabcolsep}{3pt}
\setlength{\aboverulesep}{0pt}\setlength{\belowrulesep}{0pt}\renewcommand{\arraystretch}{1.12}
\begin{tabularx}{\textwidth}{@{}l YYYYY@{}}
\toprule
Analyst input & Camera motion & Camera--camera & Camera--object & Camera--region & Camera, all four \\
Questions & 74 & 93 & 86 & 83 & 336 \\
\midrule
\rowcolor{tabgray}
Photo only, no analyst & 36.5 & 31.2 & 23.3 & 41.0 & 32.7 \\
\textbf{Latent features}, \method & 68.1\noagent{$<$10$^{-4}$} & 43.7\noagent{0.18} & 37.5\noagent{0.036} & 43.8\noagent{0.86} & 47.5\noagent{$<$10$^{-4}$} \\
\dashsep{1-6}
Question alone, no tools & 50.0\noagent{0.041} & 32.3\noagent{1.00} & 24.4\noagent{1.00} & 43.4\noagent{0.84} & 36.9\noagent{0.14} \\
Photos, no tools & 26.2\noagent{0.049} & 36.4\noagent{0.54} & 25.4\noagent{0.75} & 47.8\noagent{0.21} & 34.2\noagent{0.73} \\
Pixels in place of features & 61.6\noagent{$<$0.001} & 37.0\noagent{0.31} & 33.7\noagent{0.064} & 38.6\noagent{0.86} & 41.9\noagent{0.003} \\
\bottomrule
\end{tabularx}
\end{table*}

\Cref{tab:app_source} isolates the source of the gain of \method. The answering model and its input format stay fixed, and only what the analyst receives changes. With \emph{Question alone}, the analyst writes its report from the question and the answer options, with no features and no tools. With \emph{Photos}, it also sees the photos, still with no tools. With \emph{Pixels}, the photo-blind analyst selects the same predefined tools, which run on the RGB patches of the photos on the grids of CLIP and VGGT in place of the features, without the tools that need CLIP's text features. Photo only, \method\ and the controls are rerun here on the same GPUs with the same batch size. Some questions have no answer in three of the runs, so accuracy is over the answered questions, each test uses the questions answered under both conditions, and the $p$-values are not corrected for multiple comparisons. The gains of \method\ on camera motion and on the four camera types remain significant after a Bonferroni correction over the 20 tests of \Cref{tab:app_source}. All conclusions of this section hold when the questions without an answer are run again, so that every condition answers all 1,000 questions.

\paragraph{Latent features give the largest gain on camera questions.}
On the four camera question types, \method\ raises accuracy from $32.7\%$ to $47.5\%$, at $81$W/$36$L and $p{<}10^{-4}$, which reproduces the gain of the main text in an independent run. A report written from the question alone or from the photos does not differ significantly from photo only, on the four camera types together or over all questions ($p{\ge}0.14$). The Multi-Latent features score higher than Pixels on each of the four camera types and reach $47.5\%$ against $41.9\%$ on the four together, where Pixels also beat photo only ($p{=}0.003$). The largest gain is on camera motion, from $36.5\%$ to $68.1\%$ at $23$W/$1$L.

\paragraph{Without the ``Cannot be determined'' questions, the latent features keep the camera gain.}
In 68 questions, one option is ``Cannot be determined'' or ``None of the above'', and it is the correct answer in 49 of them. A report that calls the evidence inconclusive moves the answer toward this option, so every analyst input gains on these questions, from $33.8\%$ for photo only to between $57.4\%$ and $77.9\%$. Without them, on the remaining 308 camera questions, \method\ scores $45.1\%$ against $35.4\%$ for photo only, at $63$W/$35$L and $p{=}0.006$, whereas Pixels and Question alone do not differ significantly from photo only ($p{=}0.22$ and $p{=}1.00$). \method\ also outperforms Pixels there, $45.1\%$ against $38.1\%$ at $50$W/$30$L and $p{=}0.033$. On all 932 remaining questions, no analyst input differs significantly from photo only, and \method\ scores $35.9\%$ against $34.6\%$ with $p{=}0.48$. The gain of the Multi-Latent features is thus concentrated on camera questions, as in the main text.

\section{Further Baselines}
\label{sec:app_more}

\begin{table*}[t]
\centering
\caption{\textbf{Further baselines.} Accuracy in \% of Qwen3.8-27B on the 986 MMSI-Bench questions with a paired photo-only answer. The baselines use the photos or the camera poses decoded by VGGT. Gray numbers are the $p$-values of paired exact sign tests against photo only.}
\label{tab:app_more}
\footnotesize
\setlength{\tabcolsep}{3pt}
\setlength{\aboverulesep}{0pt}\setlength{\belowrulesep}{0pt}\renewcommand{\arraystretch}{1.12}
\begin{tabularx}{\textwidth}{@{}l YYY@{}}
\toprule
Condition & All & Camera, all four & Camera motion \\
Questions & 986 & 334 & 74 \\
\midrule
\rowcolor{tabgray}
Photo only & 33.7 & 31.4 & 37.8 \\
\textbf{\method} & 39.5\noagent{0.001} & 50.6\noagent{$<$10$^{-4}$} & 71.6\noagent{$<$10$^{-4}$} \\
\dashsep{1-4}
Self-consistency, majority of five answers from the photos & 33.9\noagent{0.94} & 32.6\noagent{0.64} & 35.1\noagent{0.77} \\
Camera poses decoded by VGGT, given as text & 35.1\noagent{0.31} & 35.6\noagent{0.098} & 51.4\noagent{0.064} \\
\bottomrule
\end{tabularx}
\end{table*}

\Cref{tab:app_more} adds baselines that use the photos or the decoded outputs of an encoder, on the 986 questions of \Cref{tab:app_stats}. Neither beats photo only significantly in any column ($p{\ge}0.064$), while \method\ does in all three columns.

\paragraph{Decoded camera poses do not replace the latent features.}
VGGT predicts a camera pose for each photo with its camera head. We convert the relative pose between the first photo and each other photo, and between consecutive photos, into sentences that state how many degrees the camera turns and tilts and in which direction it moves, and the answering VLM receives these sentences in place of the report. The decoded poses raise accuracy on the four camera types from $31.4\%$ to $35.6\%$ ($p{=}0.098$), whereas \method\ reaches $50.6\%$, at 89W/39L against the poses and $p{<}0.001$. On camera motion, the poses score $51.4\%$ and \method\ $71.6\%$, at 18W/3L and $p{=}0.001$.

\paragraph{More answers from the photos do not explain the gain.}
Self-consistency samples five answers from the photos at temperature $0.7$ and takes the majority. It scores $33.9\%$ overall ($p{=}0.94$) and $32.6\%$ on the camera types ($p{=}0.64$), and \method\ outperforms it by $5.6$~pp at 196W/141L and $p{=}0.003$. A report written from the question alone, which adds the same analysis step without features, also does not differ significantly from photo only in \Cref{tab:app_source}.

\section{Examples on Other Benchmarks}
\label{sec:app_examples_benches}

\begin{figure*}[t]
    \centering
    \includegraphics[width=0.98\textwidth]{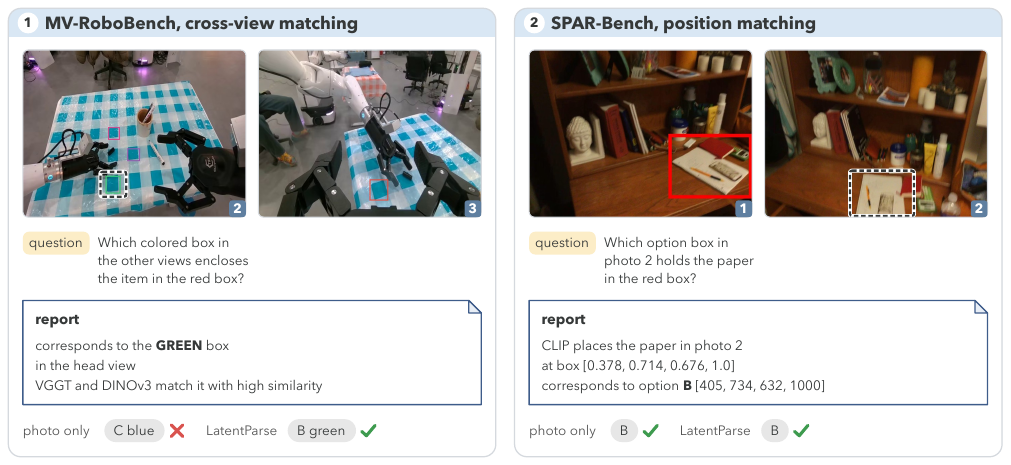}
    \caption{\mascot\ \textbf{Examples on MV-RoboBench and SPAR-Bench.} Qwen3.8-27B with the marks drawn into the photos. The dashed box is the region the analyst returns. On MV-RoboBench, \method's report points to the green box and leads to the right answer where photo only fails. On SPAR-Bench, the paper located from the features alone falls on option B. Report lines are abridged excerpts.}
    \label{fig:examples_benches}
\end{figure*}

\Cref{fig:examples_benches} shows one question from each of the other two benchmarks, with Qwen3.8-27B and the marks drawn into their photos.

\paragraph{Two encoders match the marked item across views.}
MV-RoboBench question QA091 of the cross-view matching task outlines an item with a red box in the right-gripper view and asks which colored box holds the same item in the other views. The analyst matches the red box into the head view with VGGT and DINOv3 and reads both matches as the green box. \method\ answers B, the green box, which is correct, while photo only answers C.

\paragraph{The features alone find the marked object in the second photo.}
SPAR-Bench-Tiny question 673 marks a paper with a red box in the first photo and asks which of four boxes holds it in the second. Without seeing either photo, the analyst locates the paper in the second photo with CLIP and DINOv3, and the box it reports covers option B. \method\ answers B, which is correct.

\end{document}